\documentclass[sigconf]{acmart}

\usepackage{import}
\usepackage{amsmath}
\usepackage{multirow}
\usepackage{url}

\DeclareMathOperator*{\argmax}{arg\,max}

\newcommand{\Rpos}{\ensuremath{\mathbb{R}_{\geq 0}}}
\newcommand{\x}{\ensuremath{\mathbf{x}}}

\newcommand{\y}{\ensuremath{\mathbf{y}}}

\newcommand{\EX}{\ensuremath{\mathbb{E}}}% expected value

\newcommand{\preferencegraph}{\ensuremath{R}}
\newcommand{\conflictgraph}{\ensuremath{G}}
\newcommand{\tasks}{\ensuremath{T}}
\newcommand{\rel}{\ensuremath{L}}

\newcommand{\relround}{\texttt{Relax-Round}}
\newcommand{\quadratic}{\texttt{Quadratic}}
\newcommand{\pipage}{\texttt{Pipage}}
\newcommand{\randpip}{\texttt{RPipage}}
\newcommand{\random}{\texttt{Random}}
\newcommand{\greedy}{\texttt{Greedy}}

\newcommand{\manual}
{\texttt{Manual}}
\newcommand{\GC}{\textsc{Team Formation amidst Conflicts}}
\newcommand{\shortGC}{\textsc{TFC}}
\newcommand{\shortrel}{\textsc{Relaxed-TFC}}
\newcommand{\compact}{\texttt{Compact}}
\newcommand{\linear}{\texttt{Linear}}
\newcommand{\sparsify}{\texttt{Sparsify}}
\newcommand{\concave}{\texttt{Concave}}

\newcommand{\classA}{\textit{Class-A}}  % 506
\newcommand{\classB}{\textit{Class-B}}  % 519
\newcommand{\classC}{\textit{Class-C}}  % 549
\newcommand{\classD}{\textit{Class-D}}  % 701
\newcommand{\diversity}{\textit{Company}}
\newcommand{\synth}{\textit{Synth-TF}}

\newcommand{\linnorm}{\textit{LinNorm}}
\newcommand{\inverse}{\textit{Inverse}}

\newtheorem{theorem}{Theorem}
\newtheorem{corollary}{Corollary}
\newtheorem{lemma}{Lemma}

\newtheorem{proposition}{Proposition}
\newtheorem{assumption}{Assumption}

\newtheorem{property}{Property}

\newcommand{\spara}[1]{\smallbreak\noindent{\bf{#1}}}

\newcommand{\etal}{{et al.}}

\usepackage[show]{chato-notes}
\usepackage{algorithm}
\usepackage{algpseudocode}

\newcommand*{\belowrulesepcolor}[1]{%
	\noalign{%
		\kern-\belowrulesep
		\begingroup
		\color{#1}%
		\hrule height\belowrulesep
		\endgroup
	}%
}
\newcommand*{\aboverulesepcolor}[1]{%
	\noalign{%
		\begingroup
		\color{#1}%
		\hrule height\aboverulesep
		\endgroup
		\kern-\aboverulesep
	}%
}

\newcommand{\squishlist}{\begin{list}{$\bullet$}
  { \setlength{\itemsep}{0pt}
     \setlength{\parsep}{3pt}
     \setlength{\topsep}{3pt}
     \setlength{\partopsep}{0pt}
     \setlength{\leftmargin}{1.5em}
     \setlength{\labelwidth}{1em}
     \setlength{\labelsep}{0.5em} } }
\newcommand{\squishend}{
  \end{list}  }

\AtBeginDocument{%
  \providecommand\BibTeX{{%
    \normalfont B\kern-0.5em{\scshape i\kern-0.25em b}\kern-0.8em\TeX}}}

\copyrightyear{2024} 
\acmYear{2024} 
\setcopyright{rightsretained} 
\acmConference[WWW '24]{Proceedings of the ACM Web Conference 2024}{May 13--17, 2024}{Singapore, Singapore}
\acmBooktitle{Proceedings of the ACM Web Conference 2024 (WWW '24), May 13--17, 2024, Singapore, Singapore}\acmDOI{10.1145/3589334.3645444}
\acmISBN{979-8-4007-0171-9/24/05}

\begin{document}

\title{Team Formation amidst Conflicts}

\author{Iasonas Nikolaou}
\orcid{1234-5678-9012}
\affiliation{%
  \institution{Boston University}
  \city{Boston}
  \country{USA}
}
\email{nikolaou@bu.edu}

\author{Evimaria Terzi}
\affiliation{%
  \institution{Boston University}
  \city{Boston}
  \country{USA}
}
\email{evimaria@bu.edu}

\begin{abstract}
 In this work, we formulate the problem of \emph{team formation amidst conflicts}. The goal is to assign individuals to tasks, with given capacities, taking into account individuals' task preferences and the conflicts between them. Using dependent rounding schemes as our main toolbox, we provide efficient approximation algorithms. Our framework is extremely versatile and can model many different real-world scenarios as they arise in educational settings and human-resource management. We test and deploy our algorithms on real-world datasets and we show that our algorithms find assignments that are better than those found by 
 natural baselines. In the educational setting we also show how our assignments are far better than
 those done manually by human experts. In the human-resource management application we show how our 
 assignments increase the diversity of teams. Finally, using a synthetic dataset we demonstrate 
 that our algorithms scale very well in practice.
 \end{abstract}




\maketitle

\section{Introduction}
In large project-based classes instructors often  need to 
create \emph{teams} of students and assign them to a finite number of projects they have 
available.  %Each student is assigned to one project and each project has a certain capacity, i.e., 
%only a certain number of students can be assigned to it.
%Students provide their preferences with respect to projects and teammates and their satisfaction
%with an assignment is determined by the degree to which these preferences were respected; i.e., 
Students are happy if they are in a team with friends and they work on a project  they like.
Additionally, student teams are efficient if there are no time conflicts between the team members; i.e., conflicts that stem from their class schedule.
Traditionally, such assignments are done  in an adhoc manner  or  manually by some
admin who can spend several days on the task.
%who takes into consideration students' preferences for projects and teammates; 
%in this  case it may take several days to produce an assignment.

Motivated by such applications in the education domain, we formally define the above problem as a combinatorial optimization problem. 
For this, we assume two inputs: the \emph{preference graph} and the 
\emph{conflict graph}. The former captures the preferences of users to projects; this is a bipartite graph with edge weights that are proportional to how much a student likes a project. 
The latter captures the conflicts between students, i.e., there is a (weighted) edge between two students if they are incompatible. Our goal is  to find an assignment of students to projects such that every student is assigned to one project and every project is not assigned more students than
its capacity.  The objective is to maximize the sum of the weights of the edges
of the preference graph that participate in the assignment and and the sum of the weights of
the conflict edges across the formed teams.  Students assigned to the same project form a team so we call this problem the
{\GC} ({\shortGC}) problem and we show  it is NP-hard.  %In fact, our objective 
%function does not have any discernible properties in order to be optimized efficiently.

In this paper, we present an algorithmic framework for approximating  {\shortGC}.
This framework, consists of two steps: first, our objective is replaced by
a concave objective -- which can be optimized in polynomial time and produce a fractional solution.
Then, the fractional solution
is rounded, using dependent-rounding techniques~\cite{ageev2004pipage, gandhi2002dependent}.
For our framework to work we need the original objective and its concave relaxation
to match on integral inputs.

This general framework is not new. 
In fact, it is inspired by 
the \emph{Max-$k$-Cut with given part sizes} problem~\cite{ageev2004pipage}.
In fact, our problem is identical to the \emph{Max-$k$-Cut with given part sizes}, except for the 
fact that we also have an additional linear term in our objective. We show that their 
approximation algorithm can  be applied to our problem.
Our contribution is %in a second (different) instantiation of the framework that leads to 
a much more efficient randomized algorithm with better approximation ratio on expectation. 
%Our algorithm uses a different
%concave relaxation of the original objective. 
%Additionally, it is much more efficient due to the rounding scheme it uses.

To the best of our knowledge 
the dependent-rounding techniques we use here, such as pipage and randomized pipage,
have not been widely used in practical applications; 
they primarily stem from work in theoretical computer
science~\cite{ageev2004pipage, chekuri2010dependent}. 
We see the deployment of these techniques in practice as a contribution by itself.

Using real, anonymized data, from large classes \footnote{We obtained an IRB exemption to use the anonymized version of this data.}, we demonstrate that our algorithms work extremely well in practice. 
In our experiments, we show that 
the solutions we obtain are much better compared to the manual solutions produced by a course admin across different dimensions and metrics.

Our problem formulation is  general and goes beyond educational settings.
For example, we can use our framework in human-resource management 
in order to increase the diversity of departments in companies; 
depending on the dimension across which we want to diversify we can appropriately define the 
conflict graph. %and solve the resulting instance of the {\shortGC} problem. 
In our experiments,
we show how to achieve gender diversity in a company's departments using this idea.

The generality of our framework calls for efficient algorithms. Part of our contribution is
a set of speedup techniques that allow us to apply our approximation algorithms to reasonably large 
data.  In our experimental evaluation we demonstrate that these techniques work extremely well in practice.

\spara{Discussion:} We note here that it was a design decision from our part to 
consider the conflict graph and
maximize conflicts across teams (instead of friends within teams).  We believe that this choice 
gives us greater modeling flexibility to apply our model to a variety of settings. 
For example, the conflict graph better models time conflicts among collaborators 
as well as diversity constraints.

\section{Related work}\label{sec:relatedwork}
To the best of our knowledge, we are the first to define and approximate the
{\shortGC} problem. 
However, our work is related to works in \emph{team formation}, in the data-mining literature,
as well as other 
\emph{assignment} and \emph{clustering} problems, in the theoretical computer science  literature.
We review these works below.

\spara{Team formation:} In terms of application, our work 
 belongs to the team-formation literature \cite{anagnostopoulos2010power, anagnostopoulos2018algorithms, lappas2009finding, vombatkere2023balancing, kargar2013finding, majumder2012capacitated}. 
    Most of these works consider the problem of assigning  groups of individuals to 
    tasks (one group per task) such that the tasks are completed and some objective (usually related to the well-functioning of the team or the well-being of the individuals) is optimized.
   % All these works are generalizations of the set-cover problem, or of the maximum coverage problems.
   % As a result, our objective of maximizing the overall satisfaction of the individuals is orthogonal
    %to the objectives that were used in the past. Moreover, 
    Our problem is  a partitioning problem and as such is much more complicated
    than problems of finding a good team for each available task. Additionally, in many of the existing works the objective function has a well-defined structure (e.g., is monotone and submodular, concave). 
Our objective does not have such a structure and while this gives us modeling power, optimizing it requires more advanced techniques.

  \spara{Partitioning problems:} The \emph{Max-$k$-Cut with given part sizes} 
  problem~\cite{ageev2004pipage} served as an inspiration for our model.  
  In fact, our $1/2$-approximation algorithm is a very close variant of the algorithm
  presented there. However, our objective function is slightly different from the 
  one defined by Ageev {\etal} -- due to an additional linear term. This allows us to design algorithms
  tailored to our problem, which achieve 
better approximation ratios under certain assumptions. Additionally, we focused on developing scalable
algorithms as the running time of the algorithm proposed by Ageev {\etal} was not a computationally feasible approach.

  %\note[ET]{check if the above sentence is correct.}
  
  %Moreover, a dependent randomized rounding scheme called \emph{swap rounding} \cite{chekuri2010dependent} satisfies all of the required properties and may be applied to our problem in place of randomized pipage rounding.
  %%%XXXX move this comment to the rounding
  
 \spara{Clustering problems:} One can view  our problem as a clustering problem with capacity constraints \cite{zhu2010data, negreiros2006capacitated}. Our model, however, is quite different from these works both in the objective function and the constraints.
    
% \iffalse
%     \item roundings: The main tool that we exploited in the design of our algorithm is \emph{pipage} rounding \cite{ageev2004pipage, ageev20010, chekuri2010dependent, gandhi2002dependent}. The \emph{Max-$k$-Cut with given part sizes} problem analyzed in the application section of \cite{ageev2004pipage} served as an inspiration for our model. Moreover, a dependent randomized rounding scheme called \emph{swap rounding} \cite{chekuri2010dependent} satisfies all of the required properties and may be applied to our problem in place of randomized pipage rounding.
% \fi    
    \spara{Assignment problems:}
    Our problem can be viewed as a generalization of the \emph{weighted assignment} problem \cite{kuhn1955hungarian}, where the goal is to assign individuals to tasks taking into account the task preferences of individuals. In our problem, apart from task preferences, we also have a graph capturing the relationships (or conflicts) between individuals. This additional structure increases the complexity of the problem significantly.
    Our problem is also related to the famous \emph{stable marriage}~\cite{mcvitie1971stable} problem and its variants~\cite{gale1962college, gusfield1989stable}. %In stable marriage the goal is to match entities that belong to one group to entities that belong to another. The computed matching has to be stable, i.e. it should be impossible to switch two individuals so that both of them are better off. 
    %Diverging from this line of work, 
    However, we don't look for a stable matching. Instead, our goal is to optimize an objective function capturing the overall satisfaction of individuals.
    % In our model switching two individuals may affect individuals that were not involved in the switching. This is because apart from the bipartite graph we also have a graph capturing the relationships (or conflicts) between individuals.

    \spara{The metric-labeling problem}: A minimization version of our problem is the 
    \emph{metric-labeling} problem~\cite{kleinberg2002approximation}, where the goal is to assign one of $k$ labels to each node (i.e., partition the nodes). Every assignment incurs
    assignment costs (based on the choice of label for each node) and separation costs (based on the choice of labels for "related" nodes). 
    In the capacitated version of metric-labeling~\cite{andrews2011capacitated} we are also given a capacity for each partition. 
    The main disadvantage of the algorithm developed for this version~\cite{andrews2011capacitated} is that the capacity constraints are violated by a multiplicative factor. Also, the algorithm only works for labels with uniform capacities. Finally, the approximation factor of the proposed algorithm depends on the number of labels (i.e. tasks). Defining  {\shortGC}  
    as a maximization problem allows us to overcome all of the above disadvantages.

%     \iffalse
%     \item  whole-page/page-level optimization:
%     Finally, a relevant application is whole-page \cite{devanur2016whole} or page-level \cite{lo2021page} optimization. In such problems the goal is to recommend items or show ads, while taking into consideration the relationships and/or constraints between the items presented on the same web-page (e.g. diversity constraints, redundancy between items, conflicts between advertisers). 
%     This comprises an assignment problem with extra constraints.
%     In contrast to our problem, whole-page optimization is usually considered in the online case where users arrive in an online fashion.
% \fi

\section{Problem definition}
In this section, we provide the necessary notation and we formally define the problem we solve
in this paper.

\spara{Notation:} Throughout the paper, we assume that we are given a
weighted (undirected) graph $\conflictgraph = (V, E_\conflictgraph, w)$ with
$w : E_\conflictgraph \rightarrow \Rpos$.
More specifically, 
each node $v \in V$ corresponds to an individual;
the weight $w_{uv}$ of an edge $(u, v) \in E_\conflictgraph$ captures the degree of 
conflict  between individuals $u$ and $v$. 
We call graph {\conflictgraph} the \emph{conflict graph}.

In addition to the conflict graph $G$, we also assume a preference graph {\preferencegraph}, which is 
a \emph{bipartite graph}, i.e., $\preferencegraph =(V,\tasks,E_\preferencegraph,c)$. 
The one side of the graph corresponds to individuals ($V$), the other side to 
items or tasks $\tasks$. The edges ($E_\preferencegraph$) capture the preferences of individuals to 
projects. More specifically $c: V \times \tasks \rightarrow \Rpos$ is a \emph{preference} function, where $c_{vt}$ captures the satisfaction of individual $v \in V$ when assigned to task $t\in \tasks$. Without loss of generality we assume that $0 \leq c_{ut} \leq 1$.

Throughout, we assume that each individual $v\in V$ is assigned to exactly one task and that each task
$t\in\tasks$ has \emph{capacity} $p_t$, which is task-specific.

%In this section we will provide an alternative, but useful view of our problem as a many-to-one bipartite matching with conflicts problem.
%Consider a bipartite graph $G = (A;B, E)$, where the nodes in $A$ correspond to partitions and the nodes in $B$ correspond to individuals. Every node on left part $v_t \in A$ has a capacity $p_t$ (corresponding to part $X_t$). Every node on the right part $B$ must be matched with exactly one node of $A$ (i.e. each individual must be matched with exactly one part).
%The goal is to find a valid matching maximizing the objective function.

%We have to partition $V$ into $X_1, \ldots, X_k$ under the constraints $|X_t| \leq p_t$. We assume that $\sum_{t=1}^k p_t \geq |V|$, i.e. all individuals may be assigned to a partition.
%Let $X = \{X_1, \ldots, X_k\}$.
%The goal is to partition the graph so as to maximize the total satisfaction of the students given by the following (integral) objective.
%We are also given a \emph{preference} function $c: V \times X \rightarrow \Rpos$, where $c_{vt}$ captures the satisfaction of individual $v \in V$ when assigned to partition $X_t$.

\spara{The {\GC} problem:} Given the above, our goal is to assign individuals to tasks
such that the overall satisfaction of individuals is maximized; the satisfaction of each individual is
measured by how much they like the task they are assigned to and the lack of conflicts with 
the other individuals assigned to the same task.  
We capture this intuition formally in the form of a (quadratic) program. For this, we define
 binary variables $x_{vt}$ such that $x_{vt}=1$ if individual $v$ is assigned to task $t$ and
 $x_{vt}=0$ otherwise.  Thus, our goal is the following:

\begin{align} 
    \text{max} \quad &
    F(\x) = \lambda \sum_{v \in V} \sum_{t\in \tasks} c_{vt} x_{vt} 
    + \sum_{(u,v) \in E_\conflictgraph} w_{uv} (1 - \sum_{t\in\tasks} x_{ut}x_{vt}) \label{eq:objective}\\
    \text{s.t.} \quad &
    \sum_{t\in \tasks} x_{vt} = 1, \forall v \in V \label{eq:constraint1}\\
    &\sum_{v \in V} x_{vt} \leq p_t, \forall t \in \tasks\label{eq:constraint2}\\
    &x_{vt} \in \{0,1\}, \forall v \in V, \forall t \in \tasks \label{eq:constraint3}
\end{align}
We call the problem captured by the above program {\GC} or {\shortGC} for short.
The linear term of the objective captures the satisfaction of assigning individuals to tasks and we 
call it the \emph{task satisfaction term}: $F_\preferencegraph = \sum_{v \in V} \sum_{t\in \tasks} c_{vt} x_{vt} $. 
The quadratic term captures conflicts in the following sense. The objective increases by $w_{uv}$ whenever there is conflict between individuals $u$ and $v$ and they are assigned to different tasks.
We call this term the \emph{social satisfaction term}, i.e.,
$F_\conflictgraph = \sum_{(u,v) \in E_\conflictgraph} w_{uv} (1 - \sum_{t\in\tasks} x_{ut}x_{vt})$;
this term models Max-k-Cut with given sizes of parts~\cite{ageev2004pipage}.

As far as the constraints are concerned: the first constraint enforces that every individual is assigned to exactly one task while the second constraint enforces that we assign at most $p_t$ individuals to task $t \in T$; $p_t$ is the
capacity of tasks. 
Observe that
our problem as represented above is 
a quadratic program with integer constraints and the objective function $F$ is
non-convex. 
This observation hints that the problem may be computationally hard. In fact, we have
the following result regarding the hardness of {\shortGC}:

\begin{lemma}\label{lm:nphard}
The {\GC} problem is NP-hard.
\end{lemma}
The proof stems from the fact that our problem includes the {\sc Max-Cut} problem~\cite{karp2010reducibility}, which is NP-hard.
To see this, consider the instance of our problem that has two 
tasks $\tasks = \{t_1, t_2\}$ such that $p_{t_1} = p_{t_2} = |V|$. Set $c_{vt} = 0, \forall v \in V, \forall t \in \tasks$. This is indeed an instance of the {\sc Max-Cut} problem; this observation concludes the proof of Lemma~\ref{lm:nphard}.

An interesting question is what is the value of the
hyperparameter $\lambda$ in the linear term and how one should go about setting it.
Observe that $\lambda$
balances
the relative importance of task preferences and conflicts; when $\lambda = 0$, the linear term vanishes and we only optimize for conflicts. 
As $\lambda$ grows  task preferences become dominant.
In general, tuning the hyperparameter is application specific. 
In Section~\ref{sec:lambda}, we discuss how we tune $\lambda$ in practice.

\iffalse
That said, an intuitive way to tune $\lambda$ is as follows.
Assume that $c_{vt} = 1$, if individual $v$ likes task $t$, otherwise $c_{vt} = 0$.
Let $d_{\text{avg}} = \frac{w(E)}{|V|}$, where $w(E) = \sum_{(u, v) \in E} w_{uv}$, be the average degree of an individual in the conflict graph. Then, set $\lambda$ to be a multiple of the average degree, i.e. $\lambda = \alpha d_{\text{avg}}$. Intuitively, assigning someone to a task they like is equivalent to $\alpha d_{\text{avg}}$ conflicts between individuals assigned to different tasks.
\fi

%\section{Preliminaries}
%\import{sections/}{preliminaries}

\section{Algorithms}\label{sec:algorithms}
In this section, we provide approximation algorithms for the 
{\shortGC} problem.  Our approach for solving this problem is the following: first, we will find a ``nice", i.e., \emph{concave}, relaxation of $F$, which we will call $\rel$. 
Then, we will optimize $\rel$  in the fractional domain. That is, we transform the original {\shortGC} problem described in Equations~\eqref{eq:objective}-\eqref{eq:constraint3} to the following concave program:
\begin{align} 
    \text{max} \quad &
    \rel(\x) \label{eq:relaxation}\\
    \text{s.t.} \quad &
    \sum_{t\in\tasks} x_{vt} = 1, \forall v \in V \label{eq:constraint11}\\
    &\sum_{v \in V} x_{vt} \leq p_t, \forall t \in \tasks \label{eq:constraint21}\\
    & 0\leq x_{vt}\leq 1, \forall v \in V, \forall t \in \tasks \label{eq:constraint31}
\end{align}

We call this problem {\shortrel} and it is clearly very 
similar to the {\shortGC} problem except for two important differences:
the objective $F$ %( Eq.~\eqref{eq:objective})
is substituted by its
relaxation $\rel$. Also, the 
integrality constraints %(Eq.~\eqref{eq:constraint3})
are substituted by
the corresponding fractional constraints. %(Eq.~\eqref{eq:constraint31}).
The relaxations of $F$ we propose have the following two properties, which are key for the results we propose: 

\begin{property}~\label{property:I} 
$L(\x)$ is \emph{concave}.
\end{property}

\begin{property}\label{property:matching}
For every
$\x \in \{0,1\}^{|V|\times |\tasks|}: F(\x) = \rel(\x)$. 
I.e., the original function and the relaxation agree on the integral values of $\x$.
\end{property}

Given that $\rel$ is concave and the constraints are linear, 
{\shortrel} can be solved using  \emph{gradient ascent}~\cite{boyd2004convex}.

Finally, by using appropriate rounding
techniques we transform our solution to an integral solution. 
This general algorithm, which we call {\relround} is described in Algorithm~\ref{algo:template}. 
{\relround} serves as a template for the 
approximation algorithms we develop.

We present two algorithms that use two different concave relaxations and rounding schemes, which in turn give different approximation guarantees and come with their own running-time implications.

% \note[ET]{Here you need to put the pseudocode of an algorithm that gives the high-level idea that is described in the text above.}

\begin{algorithm}[H]
\caption{{\relround}: A general approximation algorithm for the {\shortGC} problem.}\label{algo:template}
\begin{algorithmic}
\Require Objective function $F$, Rounding scheme $\Xi$
\State {\texttt{Relax}:} Given $F$ construct a concave fractional relaxation 
$L$ such that: $F(\x) = L(\x)$ for all integral $\x$.
\State {\texttt{Optimize}:} $\y^\ast = \argmax_{\y} L(\y)$
\State {\texttt{Round}:} $\x = \Xi(\y^\ast)$
\end{algorithmic}
\end{algorithm}

\iffalse
\note[ET]{These roundings have some properties that are important to be addressed here. Not any rounding scheme will work, but we do need to say what it is about the schemes that make them work in this context. I assume that they guarantee that they obtain an integral solution that has the same value as the fractional solution. Is that correct?}

Note that independently rounding the solution does not work since the integral solution has to satisfy the constraints.
\fi

\subsection{A deterministic $\frac{1}{2}$ - approximation algorithm}\label{sec:approximation1}
We start by describing a
$\frac{1}{2}$-approximation algorithm for 
the {\shortGC} problem. 
This algorithm was first introduced by Ageev {\etal}~\cite{ageev2004pipage} for the \emph{Max-$k$-Cut with given part sizes}. However, we present it here for completeness. %This algorithm also motivates
%our $\frac{3}{4}$-approximation algorithm we present next.
We call this algorithm {\pipage}, because it uses  \emph{pipage} rounding~\cite{ageev2004pipage} in order to instantiate
{\relround}.  We describe the concave relaxation $\rel_1$ and pipage rounding below.

\spara{The $\rel_1$ concave relaxation:}
\begin{equation}\label{eq:relaxation1}
        \rel_1(\x) = \lambda \sum_{v \in V} \sum_{t\in \tasks} c_{vt} x_{vt} 
    + \sum_{(u,v) \in E_\conflictgraph} w_{uv} \min \left(1, \min_t(2-x_{ut}-x_{vt}) \right).
\end{equation}
We have the following for $\rel_1$:
\begin{proposition}[\cite{ageev2004pipage}]\label{prop:properties1}
$\rel_1$ satisfies Properties~\ref{property:I} and~\ref{property:matching}.
\end{proposition}
The proof of Proposition~\ref{prop:properties1} relies on simple algebra (see \cite{ageev2004pipage}) and thus omitted. 
%
%Using $\rel_1()$ as our objective we 
%obtain a new -- relaxed-- version of the {\shortGC} problem, which we 
%call {\shortGC}-1; {\shortGC}-1 is captured by the following program:
%
%
%
%The fact that the objective function $\rel_1()$ is concave and that the constraints are linear,
%allows us to 
%constraints allows us to solve {\shortGC}-1 in 
%polynomial time using well-known algorithms for convex optimization
%such as \emph{gradient ascent}.

\spara{Pipage rounding:} pipage rounding takes a fractional solution
$\y$ of the {\shortGC} problem and transforms it into
an integral solution $\x$.  The following is a high-level description of the algorithm. For a more thorough
analysis and description of the algorithm we refer the reader to Appendix ~\ref{appendix: pipage} and the original paper~\cite{ageev2004pipage}. 

Pipage rounding is an iterative algorithm; at each iteration the current fractional solution $\y$
is transformed into 
a new solution $\y^\prime$ with smaller number of non-integral components.
Throughout, we will assume that any solution $\y$ is associated with the bipartite graph
$H_{\y}=(V, \tasks, E_{\y})$, where the nodes on the one side correspond to
individuals, the nodes on the other side to tasks and there is an edge $e(v,t)$
for every pair $(v,t)$ with $v\in V$ and $t\in\tasks$ if and only if $y_{vt}\in (0,1)$, i.e., 
$y_{vt}$ is fractional.

Let $\y$ be a current solution of the program
and $H_{\y}$ the corresponding bipartite graph.
If $H_{\y}$ contains cycles, then set $C$ to be this cycle. 
Otherwise, set $C$ to be a path whose endpoints have degree 1. Since $H_{\y}$ is bipartite, in both cases $C$ may be uniquely expressed as the union of two matchings $M_1$ and $M_2$. 
Suppose we increase all components of $\y$ corresponding to edges in $M_1$, while decreasing all components corresponding to edges in $M_2$ until some component reaches an integral value. Denote this solution by $\y_1$. Symmetrically, by decreasing the values of the variables corresponding to $M_1$ and increasing those corresponding to $M_2$, we get solution $\y_2$. 
We choose the best of these two solutions by calculating $F(\y_1)$ and $F(\y_2)$. 
We repeat the procedure until all variables are integral.

Observe that each iteration requires evaluating the objective function $F$ twice. This is a significant drawback of this algorithm, especially for large graphs, where the computation of the function is expensive.
In the next section we present a randomized version of pipage rounding that overcomes this problem.

\iffalse
Also, $\rel_1$ can be linearized using extra variables. Then, we can use linear programming to solve the problem.
Such an algorithm will provide a solution
$\tilde{\mathbf{x}}$, 
such that $\tilde{x}_{ut}\in [0,1]$ for every 
$u\in V$ and $t\in\tasks$. Then, using 
a rounding algorithm we obtain the final integral solution $\x$ such that $x_{ut}\in\{0,1\}$.
Since we use $\rel_1$ for the relaxation step and {\pipage} for rounding, we call this instantiation
of  Alg.~\ref{algo:template} {$\rel_1$-\pipage}.
\fi

\spara{Approximation guarantees:}
Following the analysis of Ageev {\etal}~\cite{ageev2004pipage} we can show that {\pipage} is an $1/2$-approximation algorithm for the {\shortGC} problem. Thus we have:

\begin{theorem}[\cite{ageev2004pipage}]\label{theorem:approx1}
The {\pipage} algorithm is an $\frac{1}{2}$-approximation algorithm for the {\shortGC} problem.
\end{theorem}
For completeness we present this proof in Appendix~\ref{appendix:approximationL1}

\spara{Running time:}
The overall complexity of {\pipage} consists of the running time of a gradient-ascent algorithm
that finds a fractional solution $\y^\ast$ to the {\shortrel} problem plus the running time of pipage rounding.
% The former is $O()$; 
The latter is
$O\left(\left( \mathcal{T}_F+ |V| + |\tasks|\right)|E_{\y^\ast}|\right)$, where $\mathcal{T}_F$ 
is the time required to evaluate the function $F$ and $E_{\y^\ast}$ is the number of fractional components of the initial solution $\y^\ast$.
This is because  each of the $E_{\y^\ast}$ steps 
of pipage rounding requires time $O\left(\mathcal{T}_F + |V|+|\tasks|\right)$ since we run a Depth-First-Search and two evaluations of $F$.

% \note[ET]{Can you add here the complexity of gradient ascent? (best you can do)?}

\subsection{Randomized $\frac{3}{4}$ - approximation algorithm}
Here, we present a
$\frac{3}{4}$-approximation algorithm for 
{\shortGC}. We call this algorithm {\randpip}, because we use the randomized pipage rounding  in order to instantiate
the {\relround} algorithm.  %We describe both the concave relaxation $\rel_2$ and randomized pipage rounding below.

\spara{The $\rel_2$ concave relaxation:}
$$
    \rel_2(\x) = 
    \lambda \sum_{v \in V}\sum_{t \in \tasks} c_{vt}x_{vt} - w(E_\conflictgraph) + \sum_{(u,v) \in E_\conflictgraph} \sum_{t \in \tasks} w_{uv} \min (1, x_{ut} + x_{vt}).
$$

For $\rel_2$ we have the following:
\begin{proposition} \label{prop:properties2}
$\rel_2$ satisfies Properties~\ref{property:I} and~\ref{property:matching}.
\end{proposition}

The proof of Proposition~\ref{prop:properties2} is given in Appendix~\ref{appendix:properties2}.

\spara{Randomized pipage rounding}
Here, we briefly present the randomized pipage scheme originally proposed by Gandhi~\cite{gandhi2002dependent}. 
Randomized pipage rounding proceeds in iterations,
just like (deterministic) pipage rounding. If $\y$ is the current fractional solution of the rounding algorithm, we calculate $\y_1$ and $\y_2$ (same as in pipage rounding) and then we probabilistically set $y^\prime$ equal to either $\y_1$ or $\y_2$. For more details we refer the reader to Appendix ~\ref{appendix: randpipage}.

\spara{Approximation guarantees:} In order to prove the $\frac{3}{4}$-approximation ratio
of {\randpip} for  {\shortGC}  we need the following
Lemma:

\begin{lemma}[\cite{chekuri2010dependent}]\label{lemma:negativecorrelation}
If we use $\Xi$ to denote the randomized pipage algorithm that rounds a fractional
 solution $\y$ to an integral solution $\x$, i.e. $\Xi(\y) = \x$, then $\Xi$ satisfies the
 following properties:
\begin{itemize}
    \item $\EX_{\Xi}[\x] = \y$
    \item $\EX_{\Xi}[(1-x_{ut}) (1-x_{vt})] \leq (1-y_{ut})(1-y_{vt})$, for all $u, v \in V$ and $t \in \tasks$
    % (needs explanation, for all $i,j$ incident to project or student)
\end{itemize}
\end{lemma}
The proof of this lemma is due to Chekuri {\etal}~\cite{chekuri2010dependent}, and thus omitted.
The most important consequence of Lemma~\ref{lemma:negativecorrelation} is the following proposition, the proof of which is given in Appendix~\ref{appendix:expectation}

\begin{proposition}\label{proposition:expectation}
    Under Assumption ~\ref{asm1}, for all $\x, \y$ such that $\x = \Xi(\y)$ and $\Xi$ being the randomized pipage rounding, we have that:
    $$
        \EX_{\Xi}[L(\x)] \geq \frac{3}{4} L(\y)
    $$
\end{proposition}

Now,
let
$\y^\ast$ 
be the optimal fractional solution of the {\shortrel} problem with objective $\rel_2$
and  $\Xi(\y^\ast) = \x^\ast$, with $\Xi$ being the randomized pipage rounding scheme. Also, let $\x_{\text{int}}$ be the optimal solution of the integral problem {\shortGC}.
Then, it holds that: 
\[
F(\x^\ast) = \rel_2(\x^\ast) \geq \frac{3}{4} \rel_2(\y^\ast) \geq \frac{3}{4} F(\x_{\text{int}}).
\] Thus, we have the following theorem:

\begin{theorem}
\label{theorem:approx2}
Under Assumption ~\ref{asm1}, {\randpip} is a $\frac{3}{4}$-randomized approximation algorithm for the {\shortGC} problem.
\end{theorem}

\spara{Running time:}
The overall complexity of {\randpip} consists of the running time of a gradient-ascent algorithm
that finds a fractional solution $\y^\ast$ to the {\shortrel} problem with objective
$\rel_2$ plus the running time of the randomized pipage rounding scheme, which is
$O((|\tasks| + |V|) |\tasks||V|)$;  assuming that the number of tasks $|\tasks| < |V|$, this becomes $O(|\tasks||V|^2)$. In contrast to deterministic pipage rounding, observe that randomized pipage rounding does not require evaluating the objective function. This results in a significant computational speed-up.

\spara{Discussion:} In the future, it would be interesting to examine if \emph{swap rounding}~\cite{chekuri2010dependent}, can be used in place of randomized pipage rounding and whether such a scheme can lead to more efficient algorithms.
We leave this as an open problem.

% \note[ET]{Talk about the difference in the complexity of pipage vs randomized pipage.}
% \note[ET]{Does this running time get affected by the number of edges in the conflict or the preference graph?  We should discuss this.}

\subsection{Tuning the hyperparameter $\lambda$}\label{sec:lambda}
In order for Theorem~\ref{theorem:approx2} to hold, we need to make the following assumption:
\begin{assumption} \label{asm1}
    (Balancing Assumption)
    Consider a feasible fractional solution $\y$.
    We assume that the following holds:
    \begin{align*}
        \lambda \sum_{v \in V}\sum_{t \in \tasks} c_{vt}y_{vt} - w(E_\conflictgraph)
        \geq 
        0 \\
        \lambda \geq \frac{w(E_\conflictgraph)}{\sum_{v \in V}\sum_{t \in \tasks} c_{vt}y_{vt}},
    \end{align*}
    where $w(E_\conflictgraph) = \sum_{(u,v) \in E_\conflictgraph} w_{uv}$.
    If we also assume that $0 \leq c_{vt} \leq 1$ for all $v\in V$ and $t \in \tasks$, then we have
    \begin{align*}
        \lambda 
        \geq 
        \frac{w(E_\conflictgraph)}{|V|} = \frac{d_{\text{avg}}}{2},
    \end{align*}
    where $d_{\text{avg}}$ is the average degree of the nodes in the conflict graph $G$ and we used the fact that $\sum_{t \in \tasks} y_{vt} = 1, \forall v \in V$.
\end{assumption}

The above assumption provides a way to tune the balancing parameter $\lambda$. In practice, we do the following: we introduce the \emph{balancing factor} $\alpha\in\mathbb{R}_{>0}$ and we set $\lambda$ to be 
$\lambda = \alpha\times \frac{d_{\text{avg}}}{2}$.  
% From the above discussion, if $\alpha\geq 1$, then the balancing assumption holds and so does the approximation guarantee of
% $\rel_2$-{\randpip}.
In practice, we tune $\alpha$ as follows: for
different values of $\alpha$  we evaluate the task and the social satisfaction terms $\left(F_\preferencegraph^{(\alpha)},F_\conflictgraph^{(\alpha)}\right)$. Then, we pick the value of $\alpha$ that gives the desired balance between the two terms.

\iffalse
\subsection{Discussion}
There are two points worth discussing before we conclude with this section.

First, the {\randpip} 
algorithm not only comes with better approximation guarantees, but it also has a much better
running time. The reason for this is that {\randpip} does not require evaluating the function $F$ -- which takes quadratic time. On the other hand, {\pipage} requires two evaluations of $F$ per iteration, and there are at most $|V|\times |\tasks|$ iterations; this makes {\pipage} impractical even for medium-sized datasets.

Secondly, it would be interesting to examine if \emph{swap rounding}~\cite{chekuri2010dependent}, can be used in place of randomized pipage rounding and whether such a scheme can lead to more efficient algorithms.
We leave this as an open problem.
\fi

%\subsection{Selecting the hyperparameter $\lambda$}

%\section{Computational speedups}
\section{Computational speedups}\label{sec:speedups}
We discuss here a few methods we use in order to speedup our algorithms.
All heuristics we discuss here can be applied to both {\pipage} as well as
{\randpip}.

\spara{Converting convex to linear programs:}
The algorithms we developed in Section~\ref{sec:algorithms} are based on the 
fact that the  {\shortrel} problem with objective functions
$\rel_1$ and $\rel_2$ is a concave problem with linear constraints and it can be solved in 
polynomial time via an application of gradient ascent.  In fact, we show that there is a way
to rewrite the {\shortrel} problems with objectives $\rel_1$ and $\rel_2$ as linear programs, 
by adding some extra variables.

For the {\shortrel} problem with objective $\rel_1$, this can be done as follows:
first, we 
substitute the term $\min(1, \min_t (2 - x_{ut} + x_{vt}))$ with the new variable $z_{uv}$ and 
the objective becomes:
\[
\rel_1(\x) = \lambda \sum_{v \in V}\sum_{t \in                 \tasks} c_{vt}x_{vt} 
                + \sum_{(u,v) \in E_\conflictgraph} w_{uv} z_{uv}.
\]
Then, we also
add the constraints $z_{uv} \leq 1$ and $z_{uv} \leq 2 - x_{ut} - x_{vt}, t \in T$. 
The full linear program is given in Appendix~\ref{appendix:linear1}

The corresponding linearization of the $\rel_2$ objective can be done as follows:
we substitute the term $\min(1, x_{ut} + x_{vt})$ with a new variable $x_{uvt}$ 
such that:
\[
\rel_2(\x) = 
\lambda \sum_{v \in V}\sum_{t \in \tasks} c_{vt}x_{vt} - w(E_\conflictgraph)
+
\sum_{(u,v) \in E_\conflictgraph} \sum_{t \in \tasks} w_{uv} x_{uvt}.
\]
We also 
add the constraints $x_{uvt} \leq 1$ and $x_{uvt} \leq x_{ut} + x_{vt}$.
The complete linear program is given in Appendix~\ref{appendix:linear2}.

The advantage of converting the convex problems into linear is that solving a linear program 
is much more efficient than solving a convex program with linear constraints. 
% In terms of asymptotic 
% running time the difference is...
In practice, using the Gurobi solver we obtained speedups up to 500x (see Table \ref{tab:time-heuristics}).

\spara{Sparsification:}
When the task capacities are small and the conflict graph is dense, a heuristic, we named \texttt{Sparsify}, that works well in practice is randomly removing conflict edges. That is, we keep each conflict edge with a certain probability $p$. Otherwise, with probability $(1-p)$ we discard the edge.
This greatly reduces the number of terms we need to evaluate $F_\conflictgraph$ in our objective resulting in computational speedups when optimizing the fractional relaxation.

Note that when $p = 1$, we don't alter the objective. As $p$ decreases, we remove more conflict edges, resulting in computational speedups, although our fractional solution might not be optimal. Selecting $p$ is problem specific. A rule of thumb is that the denser the conflict graph, the lower we can set $p$.

% \note[ET]{You need to state here that $p=1$ is the same as the original problem. For $p<1$ the solution is less accurate but you somehow need to give some intuition of what it is that we gain.}

\spara{{\compact}:}
Our intuition, but also our real-world datasets (see Section ~\ref{sec:education-data} and Appendix ~\ref{appendix:graphexamples}), reveal that our data have the following
pattern: the complement of the conflict graph, i.e.,  the friend graph, consists of relatively small densely-connected communities with similar task preferences. Intuitively, 
for two individuals $u,v$ that belong in the same community and have similar preferences 
we would expect that the vectors of $x_{ut}$'s and $x_{vt}$'s will be similar for all $t\in\tasks$.
Taking this to the extreme: individuals $u,v$ with the same neighbors in the conflict graph $G_\conflictgraph$ and the same preferences for tasks in $\tasks$ should have identical values 
$x_{ut}$, $x_{vt}$.

Formally, this is captured in the following theorem, which is proved in Appendix~\ref{sec:appendixcompact}:
%\begin{definition}
%Two individuals $u$ and $v$ are \emph{symmetrical} individuals if $(u, w) \in E \Leftrightarrow (v, w) \in E$.
%\end{definition}

%Then, we have the following theorem.
\begin{theorem}\label{thm:compact}
Consider two individuals $u, v \in V$ which have identical neighbors in $G_\conflictgraph$ (i.e.,$(u, w) \in E_\conflictgraph \Leftrightarrow (v, w) \in E_\conflictgraph$) and have identical project preferences (i.e., $c_{ut} = c_{vt}, \forall t \in \tasks$). Then, there exists an optimal solution $\y$ of {\shortrel} such that $x_{ut} = x_{vt}, \forall t \in \tasks$.
\end{theorem}

Motivated by the above theorem we define the {\compact} algorithm. On a high-level the idea is to compact densely-connected subgraphs into supernodes. 
Note that the supernodes need not be nodes that have identical neighborhood in $G$; 
after all, it may be unreasonable to assume that this will happen in practice. However, using a
graph-partitioning algorithm (e.g., spectral clustering \cite{von2007tutorial}, finding dense components \cite{charikar2000greedy}) we can partition the original set of nodes into supernodes with
similar neighborhoods.
Let $S$ be the set of supernodes, which is a partition of the original set of nodes $V$.  
Then, we create a conflict graph between supernodes;
the number of conflict edges between two supernodes $A$ and $B$ is approximately $|A| \times |B|$ (almost every node of $A$ is in conflict with every node of $B$). Thus, we set in this new conflict graph we set the weight of edge $(A,B)$ to be $w_{AB} = |A| \times |B|$ (assuming that each edge of the original graph has unit weight).
The next step is to solve the following \emph{compact} {\shortrel} problem:

\begin{align} %\label{compact formulation}
    \text{max} \quad &
    \rel(\x)\label{eq:L}\\
    \text{s.t.} \quad &
    \sum_{t \in \tasks} x_{vt} = 1, \forall v \in S \\
    &\sum_{v \in S} |v| x_{vt} \leq p_t, \forall t \in \tasks \\
    & 0 \leq x_{vt} \leq 1, \forall v \in S, \forall t \in \tasks. %\label{compact formulation end}
\end{align}
where we use $|v|$ to denote the number of simple nodes in the supernode $v$.

% \note[IN]{$|v|$. $S$ is supposed to be the set of supernodes.}
% \note[ET]{$|v|$ you need to explain what this is.}

Then, we unroll the solution to obtain a fractional solution for the original graph. That is, for each $v \in V$ we set $x_{vt} = x_{St}$, where $S$ is the supernode $v$ belongs to.
Finally, we round the fractional solution to obtain an integral solution. Depending on
whether we use $\rel_1$ or $\rel_2$ as our objective, we then round the fractional solution
using {\pipage} or {\randpip} respectively.

\iffalse
\begin{algorithm}[H]
\caption{{\compact}}\label{algo:compact}
\begin{algorithmic}
\State \texttt{Compact}: Partition the graph into supernodes
and define the \emph{compact} relaxation LP.
\State {\texttt{Optimize}:} $\y^\ast = \argmax_{\y} L(\y)$
\State \texttt{Unroll}: Let $\tilde{y}$ such that $\tilde{y}_{vt} = y_{St}, v \in V, t \in T$, where $S$ is the supernode $v$ belongs to.
\State {\texttt{Round}:} $\x = \Xi(\tilde{\y})$
\end{algorithmic}
\end{algorithm}
\fi

\section{Experiments}\label{sec:experiments}
In this section, we evaluate our framework using both real-world as well as synthetic datasets.
The experiments prove the effectiveness and efficiency of our algorithms as well as the versatility of our model to encompass many different real-world scenarios involving assignment problems with conflicts.
We make the code and the data we used in our experiments publicly available at \url{https://github.com/jasonNikolaou/TF-Conflicts.git}

\subsection{Baselines and Setup}

\spara{Baselines:} 
For our experiments we use the following baselines:

% \note[ET]{Which solver are you using?}
\emph{{\quadratic}:} This is the optimal algorithm, i.e., the algorithm that solves
the original {\shortGC} problem as expressed in Equations~\eqref{eq:objective}-\eqref{eq:constraint3}.
We use an off-the shelf solver to implement {\quadratic}, but even though this is a powerful solver,
we can run {\quadratic} only for small datasets since the solver is asked to optimize a non-convex quadratic function subject to integral constraints. 

\emph{{\greedy}:} The {\greedy} algorithm sequentially assigns an individual to the best team (i.e., the team that maximizes the objective function $F$) given that the constraints are satisfied. The algorithm terminates when  all individuals are assigned. We refer the reader to 
 Appendix~\ref{appendix:greedy} for a detailed analysis of {\greedy}.

\emph{{\random}:} This is an algorithm that randomly assigns each individual to a team until all individuals are assigned.

\emph{{\manual}:} This is the \emph{manual} assignment of individuals
to tasks as made by a human expert, which is available only in some datasets. 

\spara{Experimental setup:}
Our experimental setup is descried in Appendix ~\ref{appendix: experimental-setup}.
Unless otherwise stated we use the ``linearization'' speedup presented in Section~\ref{sec:speedups}.

\subsection{Datasets}
For our experiments, we use three different types of data: (a)~data of preferences of students
with respect to projects and collaborators in educational settings, (b)~data from the 
Bureau of Labor statistics that concern employees and their assignment to company departments and, finally, (c)~a synthetic dataset that we use to test the scalability of our algorithms and the
speedups described in Section~\ref{sec:speedups}. 
 We describe these datasets below. A summary of the characteristics of each dataset is shown in Table~\ref{tab:datasets}.

\begin{table}%[ht]
\small
  \caption{Number of nodes $|V|$, tasks $|T|$ and conflict edges $|E|$ for each dataset.}
  \label{tab:datasets}
  \begin{tabular}{ccccl}
    \toprule
    Dataset & $|V|$ & $|T|$ & $|E|$\\
    \midrule
     {\classB} & 28  & 7 & 359\\
    {\classC} & 26  & 6 & 311\\
    {\classA} & 168  & 14 & 13952\\
    {\classD} & 37  & 8 & 648\\
    {\diversity} & 4000 & 4 & 10166\\
    {\synth} & 1000 & 10 & 450454
\end{tabular}
\end{table}

\spara{Education data:}
\label{sec:education-data}
This is data coming from courses in a US institution~\footnote{IRB exception was obtained in order
to use an anonymized version of the data.}. In the classes we considered, there were a number of projects (with fixed capacities) available to students and each student filled in a form 
with their project preferences (each student ranked the projects from best to worst) and their preferences
with respect to other students they want to work with in the same project.
We have data from four such classes:
{\classA}, {\classB}, {\classC} and {\classD}.
These datasets do not contain a conflict graph between students, but instead a \emph{friend} graph (indicative such graphs are shown in Appendix~\ref{appendix:graphexamples}), i.e. two students that have an edge between them are friends and want to work together.
We define the conflict graph as the complement of the friend graph.
% The {\classD} conflict graph represents time conflicts between students, i.e. two students are in conflict if they don't have any shared time slots to work together.
We assign unit weight to each edge of the conflict graph.

Next, we have to construct the  preference graph by assigning weight $c_{u, p}$ for each student $u$ and project $t$.
Let $rank_u(t) \in [|\tasks|]$ be the rank of project $t$ in student's $u$ preference list (1 is the best, $|\tasks|$ is the worst).
We considered the following functions:
\squishlist
    \item inverse (\textit{Inverse}): $c_{u, t} = \frac{1}{\text{rank}_u(t)}$
    \item linear-normalized (\textit{LinNorm}): $c_{u, t} = \frac{|\tasks| - \text{rank}_u(t) + 1}{|\tasks|}$
    % \item step: $c_{u, t} = 1$, if $\text{rank}_u(t) < \frac{|\tasks|}{2}$, otherwise $c_{u, t} = 0$
\squishend
% Moreover, we can penalize the algorithm for assigning students to their least favorite projects.
% Specifically, given a threshold $l \in [k]$ and a penalty $P > 0$, for the inverse function we have $c_{u, t} = \frac{1}{rank_u(t)}$, if $rank_u(t) <= l$, otherwise $c_{u, t} = -P$. Similarly, we can alter all preference functions.

\spara{Employee data:}
Based on statistics from the \emph{U.S. Bureau of Labor Statistics}~\footnote{\url{bls.gov/cps/cpsaat11.htm}} 
for 2022, we built a dataset of employees in an company.
Specifically, we created a company with 4000 employees and four departments: IT, Sales, HR, PR. Each department has 1000 employees. IT and Sales departments are male dominated while HR and PR are female dominated. The distribution of males and females in each department are according to the data in the \emph{Management occupations} section of the \emph{U.S. Bureau of Labor Statistics}.
In our experiments, we  add conflict edges between all male employees, since our objective is 
to distribute the male employees more evenly. Equivalently, we could have added conflict edges between all females. Generally, depending on the diversity goal, one can add
conflicts judiciously to guide the diversification process. 
For employee preferences,
 we set $c_{ut} = 1$, if $t$ is the original department of $u$, otherwise $c_{ut} = 0$.
We assume that with probability $1\%$ an employee is suitable for switching to a new department. For those employees we set $c_{ut} = 1$, where $t$ is the department to which employee $u$ may switch.

\spara{Synthetic data}:
We also created a synthetic dataset ({\synth}) to test the 
speedups we discussed in Section~\ref{sec:speedups}.
For this we created $|V|=1000$ individuals. The conflict graph is defined as the complement of the following friend graph:
the friend graph is a planted partition graph where each partition has $100$ nodes connected with probability $0.99$. 
Edges across partitions are added with probability $10^{-5}$. 
For each partition we select a primary project $t$ and set $c_{vt} = 1$ for all nodes $v$ in the partition. 
Next, for each node $v$ we choose uniformly at random a project $t^\prime$ and set $c_{vt^\prime} = 1$.

\subsection{Forming teams in education settings}
In this section, we evaluate the quantitative and qualitative performance of our algorithms using the education datasets.  

\begin{figure}%[h]
  \centering
  \includegraphics[width=\linewidth]{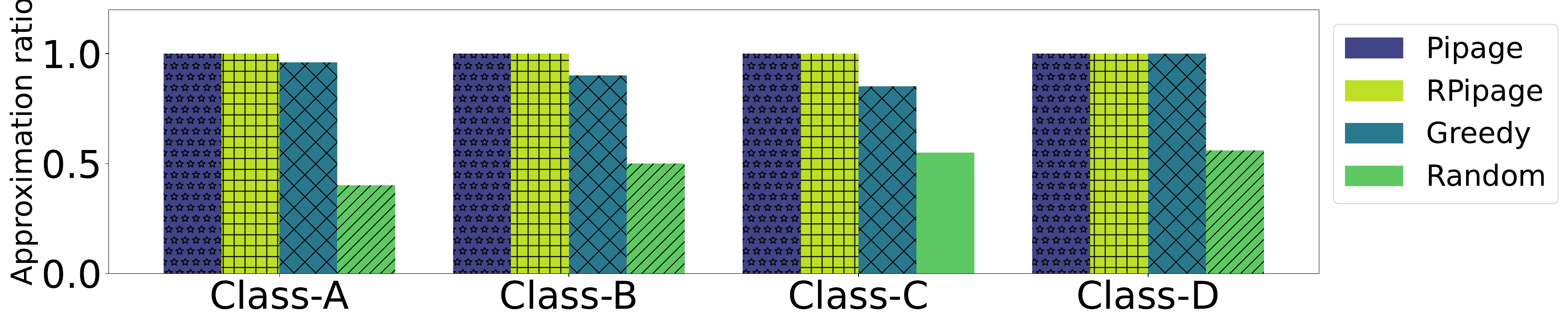}
  \caption{Education data; approximation ratio of the different algorithms. For all datasets we used the {\inverse} project-preference function and $\alpha =10$.}
  \Description{Approximation ratios}
  \label{fig:competitive-ratio-educational-inverse}
\end{figure}
\vspace{0.2cm}

\spara{Quantitative performance of our algorithms:}
We evaluate the qualitative performance of our algorithms using the \emph{approximation ratio} $\text{AR}$, i.e. the ratio of the objective function evaluated at the solution and the optimal value.
% For an algorithm $\mathcal{A}$, we define the approximation ratio of $\mathcal{A}$
% to be the ratio of the value of the objective function achieved by the solution of $\mathcal{A}$ ($\x_{\mathcal{A}}$), 
% compared to the optimal value of the objective function, i.e., $F^\ast$. The latter is computed in practice by the {\quadratic} algorithm. That is:
% \begin{equation}\label{eq:competitiveratio}
% \text{AR}(\mathcal{A})=\frac{F(\x_{\mathcal{A}})}{F^\ast}.
% \end{equation}
% $\text{AR}(\mathcal{A})$ takes values in $[0,1]$; the closer the value of
% $\text{AR}(\mathcal{A})$ is to $1$ the better the performance of algorithm $\mathcal{A}$.

Figure~\ref{fig:competitive-ratio-educational-inverse} shows the approximation ratios achieved
by our algorithms and the different baselines. For all datasets we used the {\linnorm} project preference function. The results for the \textit{Inverse} preference function are presented in Appendix~\ref{appendix: linnorm-preference}).
The results demonstrate that our algorithms have approximation ratio very close to $1$. 

Interestingly, and despite the fact that in Appendix~\ref{appendix:greedy} we show that the 
worst-case approximation ratio of {\greedy} is unbounded, in practice {\greedy} has 
\text{AR} score very close to $1$. We conjecture that this is true due to the correlation 
between students' friendships and preferences. In fact, we believe that one can bound the approximation ratio of {\greedy} under such correlation 
patterns; we leave this as future work.
%In this particular example, the performance of {\greedy} is also quite good. However, as we show in 
%Appendix~\ref{appendix:greedy} the approximation factor of {\greedy} may be unbounded.

Somewhat surprisingly, the performance of {\random} is quite good, although not comparable to the other algorithms. This can be explained by the high density of the conflict graph (or the sparseness of the friend graph) and the fact that the capacities of projects are small relative to the number
of individuals. 
Since the conflict graph is almost a complete graph, small random teams
inevitably have large number of conflict edges between them. 
That said, our analysis demonstrates that the solution given by {\random} has poor qualitative characteristics. %and does not have any practical application.

For these experiments we used $\alpha = 10$ for all datasets, since our primary goal was to assign students to projects they like. Assigning students with their friends was a secondary objective
according to the course instructors.
We chose the value of $\alpha$ following the grid-search procedure
of Section~\ref{sec:lambda};
the details are given in Appendix~\ref{appendix:tuning-edu}.
Note that by our selection of $\alpha$, the balancing assumption holds and thus the approximation guarantees of  {\randpip}  hold.

% \begin{table}[ht]
%   \caption{Approximation ratios for educational datasets}
%   \label{tab:friends_ratios}
%   \begin{tabular}{cccl}
%     \toprule
%     Dataset & Algorithm & Approximation ratio\\
%     \midrule
%     \multirow{4}{*}{\rotatebox{90}{{\classB}}} &  \pipage & 1\\
%         & \randpip & 1\\
%         & \greedy & 0.95\\
%         & \random & 0.75\\
%     \midrule
%     \multirow{4}{*}{\rotatebox{90}{{\classC}}} &            \pipage & 0.95\\
%         & \randpip & 0.95\\
%         & \greedy & 0.95\\
%         & \random & 0.80 \\
%     \midrule
%     \multirow{4}{*}{\rotatebox{90}{{\classA}}} &            \pipage & 1\\
%         & \randpip & 1\\
%         & \greedy & 0.96\\
%         & \random & 0.49 \\
%   \bottomrule
% \end{tabular}
% \end{table}

\spara{Qualitative performance of our algorithms:}
In applications such as the assignment of students to teams, what matters is not just the value
of the objective function, but the per-student satisfaction.
In this section, we analyze the solutions provided by our algorithms and our baselines
and compare those with the \emph{manual} solution provided by a domain expert who
tries to find an empirically good assignment based on the input data.

In order to evaluate the quality of the results, we 
compute the following metrics. 
Given a solution $\x$ to our problem we 
define $r(v,\x)$ to be the $\text{rank}_v (t)$, where $x_{vt} = 1$, i.e. $r(v, \x)$ is the rank -- according to $v$ -- of the task to which $v$ was assigned. Then, we define
the $\mathcal{M}$-preference metric to be:
\[
\mathcal{M}Q_\preferencegraph(\x) = \mathcal{M}(r(v,\x) \mid v\in V).
\]
In the above equation, $\mathcal{M}$ can be substituted by $\max$ or $\textit{avg}$
and the preference metric corresponds to the maximum and average ranking of the projects
assigned to students; note that the minimum is not used as it is identical across algorithms and does not provide any insight.
Intuitively, the lower the value of $\mathcal{M}Q_\preferencegraph(\x_{\mathcal{A}})$ for a solution provided by 
an algorithm $\mathcal{A}$, the better the algorithm.

Similarly, we define $\mathcal{M}Q_\conflictgraph(\x)$ to be the $\max$ or $\textit{avg}$
number of friends (non-conflicts) assigned to students in $v$. In this case, the
larger the value $\mathcal{M}Q_\conflictgraph(\x_{\mathcal{A}})$ for a solution provided by 
an algorithm $\mathcal{A}$, the better the algorithm.

%We will provide the datasets and the code on our GitHub. Finally, note that in practice we used a variant of our model, where the feasible region is non-convex, so as to accommodate for extra constraints that the teams had to obey. This version is also included in our code.

\begin{table}
\small
  \caption{Education data; $\mathcal{M}Q_\preferencegraph(\x_\mathcal{A})$  for $\mathcal{M}=\{\textit{max}, \textit{avg}\}$ of assigned project preferences and $\mathcal{A}$ being all 
  algorithms; we also report $\textit{std}$ - the standard deviation for $\textit{avg}Q_
  \preferencegraph$; {\inverse} project preference function and $\alpha = 10$.}
  \label{tab:stats_projects_inverse}
  \begin{tabular}{ccccl}
    \toprule
    Dataset & Algorithm & $\textit{max}Q_\preferencegraph$ & $\textit{avg}Q_\preferencegraph$ & $\textit{std}$\\
    \midrule
    \multirow{6}{*}{\rotatebox{90}{{\classA}}} &           \quadratic & 14.0 & 1.80 & 2.45 \\
            & \pipage & 14.0 & 1.82 & 2.51 \\
           & \randpip & 14.0 & 1.81 & 2.50 \\
           & \greedy & 14.0 & 1.95 & 2.50 \\
           & \random & 14.0 & 9.09 & 4.29 \\
           & \manual & 14.0 & 2.79 & 2.44 \\
    \midrule
    \multirow{6}{*}{\rotatebox{90}{{\classB}}} &            \quadratic & 4.0 & 1.57 & 0.90 \\
            & {\pipage} & 4.0 & 1.57 & 0.90 \\
            & {\randpip} & 4.0 & 1.57 & 0.90 \\
            & {\greedy} & 7.0 & 2.18 & 1.79 \\
            & {\random} & 9.0 & 4.75 & 2.76 \\
            & {\manual} & 3.0 & 1.71 & 0.80 \\
    \midrule
    \multirow{6}{*}{\rotatebox{90}{{\classC}}} &           \quadratic & 6.0 & 1.62 & 1.15 \\
        & \pipage & 6.0 & 1.62 & 1.15 \\
        & \randpip & 6.0 & 1.62 & 1.15 \\
        & \greedy & 6.0 & 2.12 & 1.48 \\
        & \random & 6.0 & 3.42 & 1.67 \\
        & \manual & 6.0 & 2.04 & 1.09 \\
    \midrule
    \multirow{6}{*}{\rotatebox{90}{{\classD}}} &            \quadratic & 2.0 & 1.05 & 0.23 \\
        & \pipage & 2.0 & 1.06 & 0.23 \\
        & \randpip & 2.0 & 1.06 & 0.23 \\
        & \greedy & 2.0 & 1.05 & 0.23 \\
        & \random & 8.0 & 4.45 & 2.29 \\
  \bottomrule
\end{tabular}
\end{table}

\begin{table}
\small
  \caption{Education data; $\mathcal{M}Q_\conflictgraph(\x_\mathcal{A})$  for $\mathcal{M}=\{\textit{max}, \textit{avg}\}$ of number of friends per student and $\mathcal{A}$ being all 
  algorithms; we also report $\textit{std}$ - the standard deviation for $\textit{avg}Q_\conflictgraph$; {\inverse} project preference function and $\alpha = 10$.}
  \label{tab:stats_friends_inverse}
  \begin{tabular}{ccccl}
    \toprule
    Dataset & Algorithm & ${\max}Q_\conflictgraph$ & $\textit{avg}Q_\conflictgraph$ & $\textit{std}$\\
    \midrule
    \multirow{6}{*}{\rotatebox{90}{{\classA}}} &          \quadratic &  4 & 0.65 & 0.78 \\
          & \pipage & 2 & 0.1 & 0.34 \\
          & \randpip & 1 & 0.08 & 0.26 \\
          & \greedy & 3 & 0.65 & 0.74 \\
          & \random & 1 & 0.4 & 0.49 \\
          & \manual & 3 & 0.63 & 0.63 \\
    \midrule
    \multirow{6}{*}{\rotatebox{90}{{\classB}}} &            {\quadratic} & 2 & 0.64 & 0.77 \\
            & {\pipage} & 2 & 0.64 & 0.77 \\
            & {\randpip} & 2 & 0.57 & 0.78 \\
            & {\greedy} & 3 & 0.79 & 1.08 \\
            & {\random} & 2 & 1.5 & 0.87 \\
            & {\manual} & 2 & 0.64 & 0.67 \\
    \midrule
    \multirow{6}{*}{\rotatebox{90}{{\classC}}} &          \quadratic & 2 & 0.54 & 0.57 \\
           & \pipage & 1 & 0.38 & 0.49 \\
           & \randpip & 1 & 0.38 & 0.49 \\
           & \greedy & 2 & 0.69 & 0.67 \\
           & \random & 1 & 0.54 & 0.5 \\
           & \manual & 2 & 0.69 & 0.54 \\
    \midrule
    \multirow{6}{*}{\rotatebox{90}{{\classD}}} &           \quadratic & 3 & 0.43 & 0.72 \\
        & \pipage & 1 & 0.29 & 0.45 \\
        & \randpip & 1 & 0.23 & 0.42 \\
        & \greedy & 2 & 0.38 & 0.54 \\
        & \random & 0 & 0.0 & 0.0 \\
  \bottomrule
\end{tabular}
\end{table}

Table ~\ref{tab:stats_projects_inverse} shows that students got a better project on average when using the {\quadratic}, {\pipage} and {\randpip} algorithms compared to the {\manual} assignment. Table ~\ref{tab:stats_friends_inverse} shows that students got the same (or almost the same) average number of friends using the {\quadratic}, {\pipage} and {\randpip} algorithms as in the manual assignment.
% Moreover, our analysis (see Table ~\ref{tab:stats_friends_inverse} in Appendix ~\ref{appendix: inverse-friends}) shows that students got the same (or almost the same) average number of friends using the {\quadratic}, {\pipage} and {\randpip} algorithms as in the manual assignment. %In the tables we do not provide the minimum values since they were identical for all algorithms and didn't provide any insight.

\subsection{Forming teams of employees}
In this section, we use the {\diversity} dataset in order to evaluate our algorithms' ability
to form diverse teams of employees. %in a company setting.
%Figure~\ref{fig:competitive-ratio-diversity} shows the competitive ratios of the different
%algorithms for the {\diversity} dataset.
%Our main focus for using this dataset was to demonstrate how changing the hyperparameter $\alpha$ allows us to control qualitative aspects of the solution. Specifically, we will balance the trade-off between the number of people who change department and the male-female gap per department (see Appendix ~\ref{appendix:tuning-employee}).  
%The plot that shows the trade-off 
%between $F_\conflictgraph$ and $F_\preferencegraph$ for the different values of $\alpha$ is shown in Appendix~\ref{appendix:tuning-employee}.

\begin{figure}[h]
  \centering
  \includegraphics[width=\linewidth]{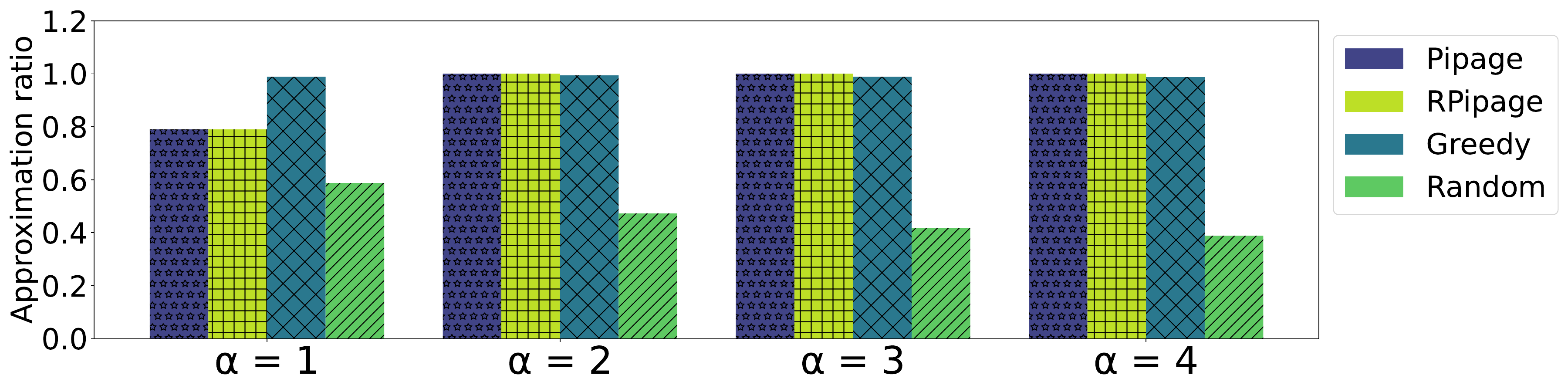}
  \caption{Employee data; approximation ratio of the different algorithms for $\alpha = 1, 2, 3, 4$.}
  \Description{Approximation ratios}
  \label{fig:competitive-ratio-diversity}
\end{figure}
\vspace{-0.5cm}

\spara{Quantitative performance of our algorithms:}
Figure ~\ref{fig:competitive-ratio-diversity} shows the 
approximation ratios of our algorithms and baselines for the {\diversity} dataset and for $\alpha = 1, 2, 3, 4$.
Somewhat surprisingly, for $\alpha = 1$ our algorithms have a approximation ratio close to $0.8$, while {\greedy} is almost optimal. 
{\random} has significantly lower approximation ratio than the other algorithms;
most of the times less than $0.5$.
In general, all other algorithms have approximation ratio close to $1$.
It is interesting to observe that in this dataset {\greedy} performs almost optimally for all choices of $\alpha$; despite the fact
that its worst case approximation factor is unbounded (see Appendix~\ref{appendix:greedy}).
As we discussed before, we believe that the reason for this  
is the correlation between conflicts and task preferences.
\iffalse
in this case this is true
by the
is the underlying patterns in the conflict graph and the correlation between conflicts and task preferences. Specifically, from the view of the friend graph, female employees constitute a large group of friends. This group can further be divided into groups that prefer a specific department (task).
\fi

\begin{figure}[h]
  \centering
\includegraphics[width=\linewidth]{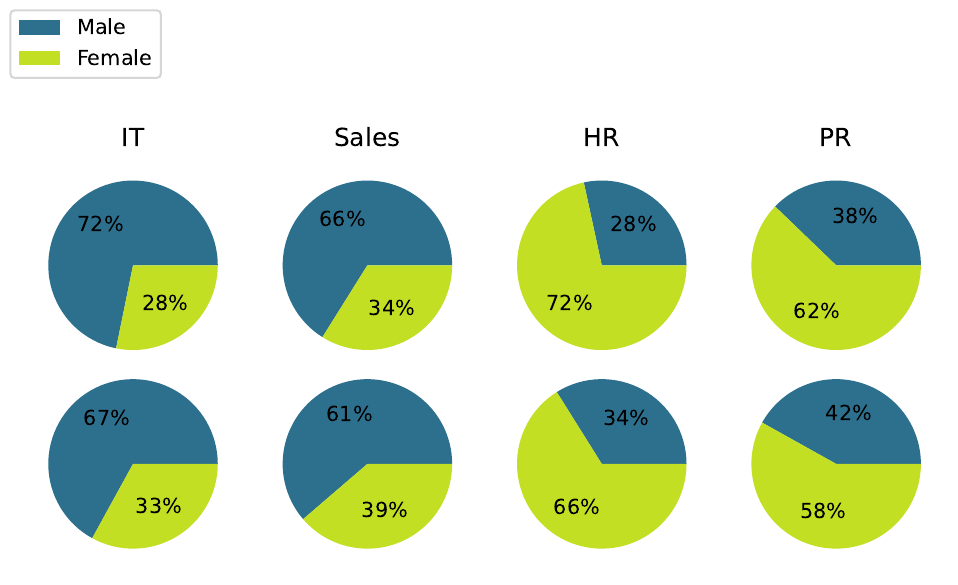}
  \caption{Employee data; Diversity per department before (1$^{\text{st}}$ row) and after (2$^{\text{nd}}$ row) we run the {\quadratic} algorithm ($\alpha = 2$).
   $8\%$ of the employees changed department. The average male-female percentage gap decreased from $35\%$ to $26\%$.}
  \Description{Percentage of male vs female employees per department before and after we run our algorithm.}
  \label{fig:diversity-before-after}  
\end{figure}

\spara{Qualitative performance of our algorithms:}
In this dataset, we evaluate the qualitative performance of our framework by showing how the optimal 
solution to our problem (obtained by {\quadratic}), affected the diversification of the
teams. Figure~\ref{fig:diversity-before-after} demonstrates exactly this. 
While only $8\%$ of the employees changed department, the average male-female percentage gap decreased from $35\%$ to $26\%$. 
Varying the value of $\alpha$ we can control this balance. Specifically, if we decrease the value of $\alpha$ the number of people who change department increases, while the average male-female gap decreases.
Hyperparameter tuning for the {\diversity} dataset is further discussed in Appendix ~\ref{appendix:tuning-employee}.

% \begin{table}
%   \caption{Approximation ratios of different algorithms for the diversity dataset}
%   \label{tab:diversity_ratio}
%   \begin{tabular}{cccl}
%     \toprule
%     Dataset & Algorithm & Approximation ratio\\
%     \midrule
%     \multirow{4}{*}{\rotatebox{90}{\textit{Diversity}}} &   \pipage & -\\
%         & \randpip & -\\
%         & \greedy & -\\
%         & \random & -
% \end{tabular}
% \end{table}

\subsection{Evaluating the speedup techniques}\label{sec:experiment-speedups}
In this section, we use the {\synth} dataset to demonstrate the speedups obtained by the different
techniques we discussed in Section~\ref{sec:speedups}. As we discussed, solving the {\shortGC} problem
as described in Equations~\eqref{eq:relaxation}-\eqref{eq:constraint31} using 
a convex solver does not scale up. Thus, we apply to this original problem sparsification and then
run the convex solver; we call this algorithm {\sparsify}-{\concave}. For the {\sparsify} algorithm we used $p = 0.01$, i.e., we kept only $1\%$ of the edges of the conflict graph.
Alternatively, we transform the problem into a problem with a linear objective by adding auxiliary variables and constraints (as discussed in Section~\ref{sec:speedups}) and run the {\randpip} algorithm.
We call this algorithm {\linear}.
Another algorithm we use is {\sparsify}-{\linear}, which combines
sparsification and linearization.  Finally, we also combine {\compact} with {\linear} to obtain the
{\compact}-{\linear} algorithm. 
    Note that for the implementation of {\compact} we use the spectral clustering algorithm (\cite{von2007tutorial}) available in scikit-learn ~\footnote{\url{scikit-learn.org/stable/modules/generated/sklearn.cluster.SpectralClustering.html}}.

\begin{figure}[h]
  \centering
  \includegraphics[width=\linewidth]{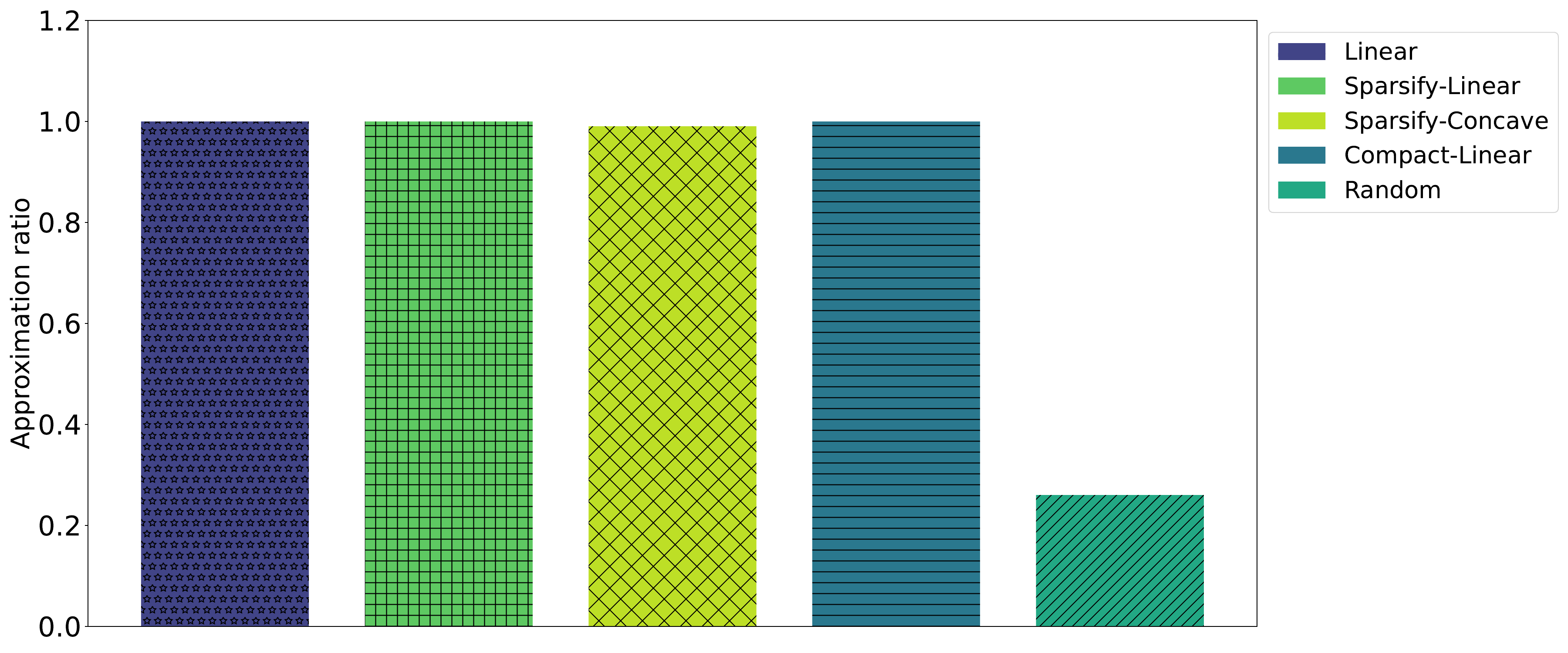}
  \caption{{\synth} dataset; approximation ratios of the speedups; We used $\alpha = 10$.}
  \Description{Approximation ratios of heuristics}
  \label{fig:competitive-ratio-heuristics}
\end{figure}

Figure~\ref{fig:competitive-ratio-heuristics} shows the approximation ratios of the above 
heuristics and Table~\ref{tab:time-heuristics} the running times of the same heuristics on the {\synth} dataset. For this experiment, we set the value of $\alpha = 10$.

Although all algorithms perform almost optimally,
speed-ups vary.
First, trying to directly optimize the {\concave} relaxation results in a time-out of our solver. Using {\sparsify} before optimizing the {\concave} relaxation renders the problem solvable in $1000$ seconds. 
Using the {\linear} algorithm yields an extra 3x speedup. 
Finally, combining {\linear} with either {\sparsify} or {\compact} further reduces the running time down to $2-3$ seconds which is a ~100x speedup. 
In total, we managed to reduce the time from $1095$ seconds using {\sparsify}-{\concave} to $2-3$ seconds combining {\linear} with one of {\sparsify} or {\compact}.

\begin{table}
\small
  \caption{Running time (seconds) for the speed-ups}
  \label{tab:time-heuristics}
  \begin{tabular}{ccl}
    \toprule
    Algorithm & Time (seconds)\\
    \midrule
    {\concave} & \text{time out} \\
    {\sparsify}-{\concave} & 1095 \\
    {\linear} &  342 \\
    {\sparsify}-{\linear} & 2 \\
    {\compact}-{\linear} & 3 \\
  \bottomrule
\end{tabular}
\end{table}

\vspace*{-0.5cm}

\section{Conclusions}
Motivated by the need to form teams of students in large project-based classes we defined
the {\shortGC} problem and showed that 
(a) it is NP-hard and that (b) it is closely related to Max-$k$-Cut with given part sizes~\cite{ageev2004pipage}. For {\shortGC}, we designed a new efficient randomized approximation algorithm and practical methods for speeding it up.
We applied our algorithms to real-world datasets and demonstrated their 
efficacy across different dimensions.  
%We believe that our contribution lies in the model definition and its application in practice, but also
%on the introduction of dependent-rounding techniques (such as pipage and randomized pipage) 
%in the data-mining community.  
In the future, we want to further explore possible speedups for our algorithm and also formally investigate the extremely good performance of {\greedy} in practice -- despite its unbounded worst-case approximation ratio.

\iffalse
We believe that our contribution lies in the application 
of the dependent-rounding techniques (e.g., pipage) in a practical setup.
Although our experiments were done in datasets of a few thousand individuals, in the future we want to explore how to further speedup {\randpip} and also investigate the relationship between the performance of greedy and the correlation between conflicts and task preferences in the input.
\fi

\iffalse
First this problem is  generalization of the 
problem of Max-$k$-Cut with given part sizes~\cite{ageev2004pipage}
(swap rounding)
(accelerating pipage rounding, can the complexity be reduced? Parallel implementation)
(greedy algorithm performance in practice)
(security our framework, can a student trick the algorithm?)\fi

\spara{Acknowledgements}
This research was supported by NSF grants 1813406 and 1908510. Evimaria Terzi was also supported by a grant from 
Max Planck Institute for Software Systems (MPI-SWS).

\clearpage
\bibliographystyle{ACM-Reference-Format}
\bibliography{references}

\clearpage

\appendix
\section{Proofs}

\subsection{Proof of Theorem~\ref{theorem:approx1}}\label{appendix:approximationL1}

We begin with the following lemma:
\begin{lemma} \label{lemma:inequality}
Let $\y$ be a (fractional) feasible solution to the {\shortGC}-1 problem. Then, for every 
$u,v\in V$ we have:
    \begin{align*}
        1 - \sum_{t \in \tasks} y_{uv} y_{vt} \geq 
        \frac{1}{2} \min \left(1, \min_{t \in \tasks} (2 - y_{uv} - y_{vt}) \right)
    \end{align*}
\end{lemma}

\begin{proof}
Let $z_{uv} = \min \left(1, \min_t (2 - y_{uv} - y_{vt})\right)$.
Define $$t^\prime=\arg\max_{t\in\tasks} (y_{ut} + y_{vt})$$ and
\begin{align} \label{def:q_uv}
    q_{uv} = y_{ut^\prime} + y_{vt^\prime}.
\end{align}
Then, 
\begin{align} \label{def:z_uv}
    z_{uv} = \min (1, 2 - q_{uv}).
\end{align}
Also, it holds that
\begin{align} \label{lemma:fact}
    \sum_{t \in \tasks} y_{ut} + y_{vt} = \sum_{t \in \tasks} y_{ut} + \sum_{t \in \tasks} y_{vt} = 1 + 1 = 2
\end{align}
as well as the arithmetic-geometric mean inequality which says that for any two 
positive numbers $a,b$:
\begin{align} \label{ineq:arith-geom}
    \left(\frac{a+b}{2}\right)^2 \geq ab. 
\end{align}
Thus, we have:
\begin{align*}
    1 - \sum_{t \in \tasks} y_{ut} y_{vt} &= 1 - y_{ut^\prime}y_{vt^\prime} - \sum_{t \neq t^\prime} y_{ut}y_{vt} \\
    &\geq 1 - \left(\frac{y_{ut^\prime} + y_{vt^\prime}}{2}\right)^2
        - \sum_{t \neq t^\prime} \left(\frac{y_{ut} + y_{vt}}{2}\right)^2 && (\text{using }\eqref{ineq:arith-geom})\\
    &\geq 1 - \left(\frac{q_{uv}}{2} \right)^2
            - \left( \frac{\sum_{t \neq t^\prime} y_{ut} + y_{vt}}{2} \right)^2
            && \hspace{-0.8cm}(\eqref{def:q_uv} \text{ and convexity})\\
    &= 1 - \left(\frac{q_{uv}}{2} \right)^2 -
    \left( \frac{2 - (y_{ut^\prime} + y_{vt^\prime})}{2} \right)^2  && (\text{using } \eqref{lemma:fact})\\
    &= 1 - \left(\frac{q_{uv}}{2} \right)^2 - \left(1 - \frac{q_{uv}}{2}\right)^2 && (\text{using } \eqref{def:q_uv})\\
    &= q_{uv} - \frac{q_{uv}^2}{2}.
\end{align*}

\emph{Case 1.} $1 \leq q_{uv} \leq 2$. Then, by Eq.~\eqref{def:z_uv}, $z_{uv} = 2 - q_{uv}$, and 
\begin{align}
    1 - \sum_{t \in \tasks} y_{ut}y_{vt} \geq q_{uv} - \frac{q_{uv}^2}{2} \geq \frac{1}{2} z_{uv}.
\end{align}

\emph{Case 2.} $0 \leq q_{uv} \leq 1$. Then, by Eq.~\eqref{def:z_uv}, $z_{uv} = 1$. By the assumption of this case, for every $t$ it holds that 
\begin{align} \label{case2}
    0 \leq y_{ut} + y_{vt} \leq 1.
\end{align}
Using the arithmetic-geometric mean inequality (Eq.~\eqref{ineq:arith-geom}), we have
\begin{align*}
    1 - \sum_{t \in \tasks} y_{ut}y_{vt} 
    &\geq 1 - \sum_{t \in \tasks} \left( \frac{y_{ut} + y_{vt}}{2}\right)^2 \\
    &= 1 - \frac{1}{4} \sum_{t \in \tasks} (y_{ut} + y_{vt})^2\\
    &\geq 1 - \frac{1}{4} \sum_{t \in \tasks} (y_{ut} + y_{vt}) && (\text{using } \eqref{lemma:fact})\\
    &= 1 - \frac{1}{4} 2\\
    &= \frac{1}{2} z_{uv}
\end{align*}

\end{proof}

A corollary of Lemma~\ref{lemma:inequality} is the following:
\begin{corollary} \label{corollary:FL-bound}
If $\y$ is a (fractional) feasible solution to problem {\shortrel} with objective $\rel_1$, then
    \begin{align*}
        F(\y) \geq \frac{1}{2} \rel_1(\y).
    \end{align*}
\end{corollary}

Let $F^\ast$ be the value of the optimal (integral) solution to  {\shortGC}. If $\y^\ast$ is the optimal (fractional) solution to {\shortrel} with objective $\rel_1$, it holds that $\rel_1(\y^\ast) \geq F^\ast$.
From Corollary \ref{corollary:FL-bound} we have that $F(\y^\ast) \geq \frac{1}{2} \rel_1(\y^\ast)$. Using pipage rounding we can round the fractional solution $\y^\ast$ to an integral solution $\bar{\x}$ such that $F(\bar{\x}) \geq F(\y^\ast)$. Thus, 

$$F(\bar{\x}) \geq F(\y^\ast) \geq \frac{1}{2} \rel_1(\y^\ast) \geq \frac{1}{2} F^\ast.$$ 

\spara{$\epsilon$-convexity property.} In order for $F(\bar{\x}) \geq F(\y^*)$ to hold, the \emph{$\epsilon$-convexity} property has to hold (see \cite{ageev2004pipage}). 
Namely, the function $\phi(\epsilon, \y, R) = F(\y(\epsilon, R))$ has to be convex in $\epsilon$ for every feasible solution $\y$ and for all paths and cycles $R$ of the graph $H_{\x}$.
We refer the reader to Section 4 in \cite{ageev2004pipage} for a formal explanation. Note that in our objective function except for the quadratic term, we also have a linear term. However, the quadratic term dominates the linear
term and $\phi(\epsilon)$ remains a  quadratic polynomial with a non-negative
leading coefficient.

%\subsection{Proof of Proposition~\ref{proposition:fractional}}\label{appendix:proposition-goemans}

\subsection{Proof or Proposition~\ref{prop:properties2}}
\label{appendix:properties2}
\spara{Property 1}
To prove concavity it suffices to see that $L_2(\x)$ is the sum of concave functions. Note that $\min(1, x_{ut} + x_{vt})$ is concave.

\spara{Property 2}: 
For the proof of Property 2, we need the following lemmas:
\begin{lemma} \label{lemma:integral}
    For all $x, y \in \{0, 1\}$ it holds that
    \begin{align*}
        1 - (1-x)(1-y) = x + y - xy = \min (1, x + y).
    \end{align*}
\end{lemma}
The proof of this lemma consists of checking equality for all combinations of integral values 
of $x$ and $y$.
%\begin{proof}
%    The proof consists of checking equality for all four value assignments for $x$ and $y$.
%\end{proof}

\begin{lemma}[\cite{goemans1994new}] \label{lemma:fractional}
     For all $x, y \in [0, 1]$ it holds that
    $$1 - (1-x)(1-y) \geq \frac{3}{4} \min (1, x+y).$$
\end{lemma}
%For completeness we present the proof of this proposition in Appendix~\ref{appendix:proposition-goemans}.

Now we are ready to prove the following:
\begin{proposition} \label{proposition: integral equality}
    For all integral $\x$ it holds that $F(\x) = \rel_2(\x)$.
\end{proposition}

\begin{proof}
    It holds that 
    \begin{align*}
        &\sum_{(u,v) \in E} w_{uv} (1 - \sum_{t \in \tasks} x_{ut}x_{vt}) =\\
            & w(E) - \sum_{(u,v) \in E} \sum_{t \in \tasks} w_{uv} x_{ut}x_{vt} = \\
            & w(E) + \sum_{(u,v) \in E} \sum_{t \in \tasks} w_{uv} \min (1, x_{ut} + x_{vt})\\
            &- \sum_{(u,v) \in E} \sum_{t \in \tasks} w_{uv} (x_{ut} + x_{vt}) =\\
            & w(E) + \sum_{(u,v) \in E} \sum_{t \in \tasks} w_{uv} \min (1, x_{ut} + x_{vt}) 
            - 2 w(E) = \\
            & \sum_{(u,v) \in E} \sum_{t \in \tasks} w_{uv} \min (1, x_{ut} + x_{vt}) - w(E),
    \end{align*}
    where in the second equality we used lemma \ref{lemma:integral}.

    Using the above we have that
    
    \begin{align*}
        F(\x) &= \lambda \sum_{v \in V} \sum_{t \in \tasks} c_{vt} x_{vt} 
    + \sum_{(u,v) \in E} w_{uv} (1 - \sum_{t \in \tasks} x_{ut}x_{vt}) \\
        &= \sum_{(u,v) \in E} \sum_{t \in \tasks} w_{uv} \min (1, x_{ut} + x_{vt}) + \lambda \sum_{v \in V} \sum_{t \in \tasks} c_{vt} x_{vt} - w(E) \\
        &= \rel_2(\x).
    \end{align*}
\end{proof}

\subsection{Proof of Proposition~\ref{proposition:expectation}}
\label{appendix:expectation}
\begin{proof}
    \begin{align*}
        \EX_{\Xi}[L(\x)] &= \EX_{\Xi}[\sum_{(u,v) \in E} \sum_{t \in \tasks} w_{uv} \min (1, x_{ut} + x_{vt}) \\ 
            &+ \lambda \sum_{v \in V}\sum_{t \in \tasks} c_{vt}x_{vt} - w(E)] \\
            &= \EX_{\Xi}[\sum_{(u,v) \in E} \sum_{t \in \tasks} w_{uv} ( 1 - (1 - x_{ut}) (1 - x_{vt})) \\ 
            &+ \lambda \sum_{v \in V}\sum_{t \in \tasks} c_{vt}x_{vt} - w(E)] \\
            &\geq \sum_{(u,v) \in E} \sum_{t \in \tasks} w_{uv} ( 1 - (1 - y_{ut}) (1 - y_{vt})) \\ 
            &+ \lambda \sum_{v \in V}\sum_{t \in \tasks} c_{vt}y_{vt} - w(E) \\
            &\geq \frac{3}{4} \sum_{(u,v) \in E} \sum_{t \in \tasks} w_{uv} \min (1, y_{ut} + y_{vt}) \\ 
            &+ \frac{3}{4} \left( \lambda \sum_{v \in V}\sum_{t \in \tasks} c_{vt}y_{vt} - w(E) \right)\\
            &= \frac{3}{4} L(\y)
    \end{align*}
    where in the second line we used Lemma \ref{lemma:integral}, in the third line we used the properties of the rounding operator, and in the fourth line we used Lemma \ref{lemma:fractional} and Assumption \ref{asm1}. 
\end{proof}

\subsection{Linearization of $\rel_1$}\label{appendix:linear1}
\begin{align} \label{linear relax l1}
    \text{max} \quad &
    \rel_1(\x) = \lambda \sum_{v \in V}\sum_{t \in                 \tasks} c_{vt}x_{vt} 
                + \sum_{(u,v) \in E_\conflictgraph} w_{uv} z_{uv} \\
    \text{s.t.} \quad &
    \sum_{t \in \tasks} x_{vt} = 1, \forall v \in V \\
    &\sum_{v \in V} x_{vt} \leq p_t, \forall t \in \tasks \\
    & z_{uv} \leq 1, \forall (u,v) \in E_\conflictgraph, \\
    & z_{uv} \leq 2 - x_{ut} - x_{vt}, \forall (u,v)\in E_\conflictgraph, t \in \tasks.
\end{align}

\subsection{Linearization of $\rel_2$}\label{appendix:linear2}

\begin{align} \label{linear formulation init}
    \text{max} \quad &
    L_2(\x) = \sum_{(u,v) \in E_\conflictgraph} \sum_{t \in \tasks} w_{uv} x_{uvt} + 
            \lambda \sum_{v \in V}\sum_{t \in \tasks} c_{vt}x_{vt} - w(E_\conflictgraph)\\
    \text{s.t.} \quad &
    \sum_{t \in \tasks} x_{vt} = 1, \forall v \in V \\
    &\sum_{v \in V} x_{vt} \leq p_t, \forall t \in \tasks \\
    & x_{uvt} \leq 1, \forall (u,v) \in E_\conflictgraph, t \in \tasks\\
    & x_{uvt} \leq x_{ut} + x_{vt}, \forall (u,v)\in E_\conflictgraph, t \in \tasks. \label{linear formulation end}
\end{align}

\subsection{Proof of Theorem~\ref{thm:compact}}\label{sec:appendixcompact}

\begin{proof}
    Let $\y$ be an optimal solution of the fractional relaxation of {\shortGC} such that it does not hold that $y_{ut} = y_{vt}, \forall t \in \tasks$. Since $\sum_t y_{ut} = \sum_t y_{vt} = 1$, there exist $t_1, t_2 \in \tasks$ such that $y_{ut_1} \neq y_{vt_1}$ and $y_{ut_2} \neq y_{vt_2}$.
    Let $\tilde{\y}$ be the "symmetrical" solution where we have replaced $u$ with $v$ and vice versa. That is,
    \begin{itemize}
        \item $\tilde{y}_{ut} = y_{vt}, t \in \tasks$
        \item $\tilde{y}_{vt} = y_{ut}, t \in \tasks$
        \item $\tilde{y}_{uv't} = y_{vv't}, t \in \tasks, (u, v') \in E$
        \item $\tilde{y}_{vv't} = y_{uv't}, t \in \tasks, (v, v') \in E$
    \end{itemize}
    For the last two equalities note that since nodes $u$ and $v$ are \emph{symmetrical} it holds that $(u, v') \in E \Leftrightarrow (v, v') \in E$.
    Then, $\x = (\y + \tilde{\y}) / 2$ is an optimal feasible solution where $x_{ut} = x_{vt}, \forall t \in \tasks$.
    To see that note that the constraints are linear and thus any convex combination of feasible solutions is also feasible.
    Moreover, the objective function is linear and thus any convex combination of optimal solutions is also optimal. 
    Finally, concluding our proof observe that
    \begin{align*}
        x_{ut_1} = x_{vt_1} = \frac{y_{ut_1} + y_{vt_1}}{2}
    \end{align*}
    and
    \begin{align*}
        x_{ut_2} = x_{vt_2} = \frac{y_{ut_2} + y_{vt_2}}{2}
    \end{align*}
    
\end{proof}

\section{Analysis of the {\greedy} algorithm}\label{appendix:greedy}

\spara{Unbounded approximation ratio of {\greedy}:}
We have the following result in terms of the performance of greedy with respect to our objective function:
\begin{proposition}
    {\greedy} has unbounded approximation ratio.
\end{proposition}
\begin{proof}
    Consider the following instance of our problem.
    $V = \{u, v, z\}$, with task preferences $c_{ut_1} = 1 - \epsilon$, $c_{vt_2} = \epsilon$ and all other preferences equal to 0.
    $\tasks = \{t_1, t_2\}$, with capacities $p_{t_1} = 1$ and $p_{t_2} = 2$.
    The conflict graph consists of the edge $(v, z)$ with weight $w_{vz} = \mathcal{W}$.
    For the objective function assume that $\lambda = 1$.
    Running the {\greedy} yields the assignment: $u$ assigned to $t_1$ and $v, z$ are assigned to $t_2$.
    The optimal assignment is $z$ assigned to $t_1$ and $u, v$ to $t_2$. The approximation ratio is:
    $$\text{AR} = \frac{(1 - \epsilon) + \epsilon}{\mathcal{W} + \epsilon} \leq
    \frac{1}{\mathcal{W}}.$$
    As $\mathcal{W} \rightarrow \infty$, the approximation ratio $\text{AR} \rightarrow 0$.
\end{proof}

\spara{Running time of {\greedy}:}
The complexity of {\greedy} is $O(|V|^2|T|\mathcal{T}_F)$, where $\mathcal{T}_F$ is the cost of calculating $F$.
We have $|V|$ iterations in total, since at each iteration we assign one individual to a team.
Each iteration costs $|V||T|\mathcal{T}_F$, since we calculate the change in the objective function when considering the addition of each remaining individual to each team which is not full.

\section{Dependent rounding schemes}
\subsection{Pipage rounding} \label{appendix: pipage}
In this section we give a description of the pipage rounding algorithm. For a more detailed analysis we refer the reader to the original paper ~\cite{ageev2004pipage}.
Pipage rounding is an iterative algorithm; at each iteration the current fractional solution $\y$
is transformed into 
a new solution $\y^\prime$ with smaller number of non-integral components.
Throughout, we will assume that any solution $\y$ is associated with the bipartite graph
$H_{\y}=(V, \tasks, E_{\y})$, where the nodes on the one side correspond to
individuals, the nodes on the other side to tasks and there is an edge $e(v,t)$
for every pair $(v,t)$ with $v\in V$ and $t\in\tasks$ if and only if $y_{vt}\in (0,1)$, i.e., 
$y_{vt}$ is fractional.

Let $\y$ be a current solution satisfying the constraints of the program
and $H_{\y}$ the corresponding bipartite graph.
If $H_{\y}$ contains cycles, then set $C$ to be this cycle. Otherwise, set $C$ to be a path whose endpoints have degree 1. Since $H_{\y}$ is bipartite, in both bases $C$ may be uniquely expressed as the union of two matchings $M_1$ and $M_2$. 
Given this, define a new solution
 $\y(\epsilon, C)$ as follows: 
 \squishlist
 \item if $e \in E_{\y} \setminus C$, then $y_e(\epsilon, C) = y_e$. 
 \item Otherwise, $y_e(\epsilon, C) = y_e + \epsilon, e \in M_1$ and $y_e(\epsilon, C) = y_e - \epsilon, e \in M_2$.
\squishend 
For the above, set
$$
\epsilon_1 = \min \{\epsilon > 0: 
    (\exists e \in M_1: y_e + \epsilon = 1) \lor 
    (\exists e \in M_2: y_e - \epsilon = 0)
    \}
$$
and
$$
\epsilon_2 = \min \{\epsilon > 0: 
    (\exists e \in M_1: y_e - \epsilon = 0) \lor 
    (\exists e \in M_2: y_e + \epsilon = 1)
    \}.
$$
Let $\y_1 = \y(-\epsilon_1, C)$ and $\y_2 = \y(\epsilon_2, C)$.
Set $\y^\prime = \y_1$, if $F(\y_1) > F(\y_2)$, and $\y^\prime = \y_2$ otherwise.
Note that $\y^\prime$ has smaller number of fractional components than $\y$ and, thus, {\pipage} terminates after at most $|E_{\y^\ast}|$ iterations, i.e., as many as the number of fractional values in the $\y^\ast$ vector output by the optimization algorithm. The following theorem states that $\y^\prime$ satisfies the following constraints:
\begin{theorem}[\cite{ageev2004pipage}]
    Consider performing pipage roudning starting from the fractional solution $\y$. Let $\x$ be the integral solution produced when {\pipage} terminates. Then,
    $$
    \lfloor\sum_{e(v,t) \in \delta (v)} y_{vt} \rfloor
    \leq
        \sum_{e(v,t) \in \delta (v)} x_{vt}
    \leq
    \lfloor\sum_{e(v,t) \in \delta (v)} y_{vt} \rfloor + 1,
    $$
where for every $v\in V$ $\delta(v)$ is the set of edges in the preference graph $\preferencegraph$, that are incident
to $v$.
\end{theorem}

Since all $p_t$'s are integers (see Eq.~\ref{eq:constraint2}), 
the above theorem implies that $\x$ is a feasible solution.

\subsection{Randomized pipage rounding} \label{appendix: randpipage}
Here, we briefly present the randomized pipage scheme originally proposed by Gandhi~\cite{gandhi2002dependent} adapted to our problem. Randomized pipage rounding proceeds in iterations,
just like (deterministic) pipage rounding. If $\y$ is the fractional solution at the current iteration of the rounding algorithm, we update $\y$ as follows:

If $e \in E_{\y} \setminus C$, then $y_e(\epsilon, C) = y_e$.
If $e \in C$, then $\y^\prime = \y_1$, with probability $\epsilon_2 / (\epsilon_1 + \epsilon_2)$. Otherwise, with probability $\epsilon_1 / (\epsilon_1 + \epsilon_2)$, $\y^\prime = \y_2$.
Note $C$, $\epsilon_1$, $\epsilon_2$, $\y_1$ and $\y_2$ are the same as the ones
defined in the description of pipage rounding.
As the number of fractional elements of $\y$ decrease in every iteration, randomized pipage rounding
terminates after at most $O(|E_{\y^\ast}|)$ iterations, where $\y^\ast$ is the solution to the {\shortrel}
problem with objective $\rel_1$.

\section{Experiments}

\subsection{Experimental setup}
\label{appendix: experimental-setup}
All of the experiments were run on a machine with an Intel(R) Xeon(R) Gold 6242 CPU @ 2.80GHz and 16GB memory.
All of our code is written in Python 3.6.8. For linear and quadratic optimization we used Gurobi ~\footnote{\url{https://www.gurobi.com/}}. For optimizing concave functions we used  CVXPY ~\footnote{\url{https://www.cvxpy.org/}}.
In all our experiments, unless otherwise explicitly stated, we use the ``linearization'' speedup we 
presented in Section~\ref{sec:speedups}. Combining this with the Gurobi solver we obtain reasonable 
running times for all our experiments.  We also demonstrate the effect of this linearization  in Section~\ref{sec:experiment-speedups}.
In fact, when we tried optimizing the concave relaxation with linear constraints as expressed in Equations ~\eqref{eq:relaxation}-\eqref{eq:relaxation1}, our solver could not terminate. For this concave
problem we used CVXPY as Gurobi only solves linear (and quadratic) problems.
Our code will be made publicly available.

\subsection{Indicative education datasets and their characteristics}\label{appendix:graphexamples}
\begin{figure}
  \centering
  \includegraphics[width=\linewidth]{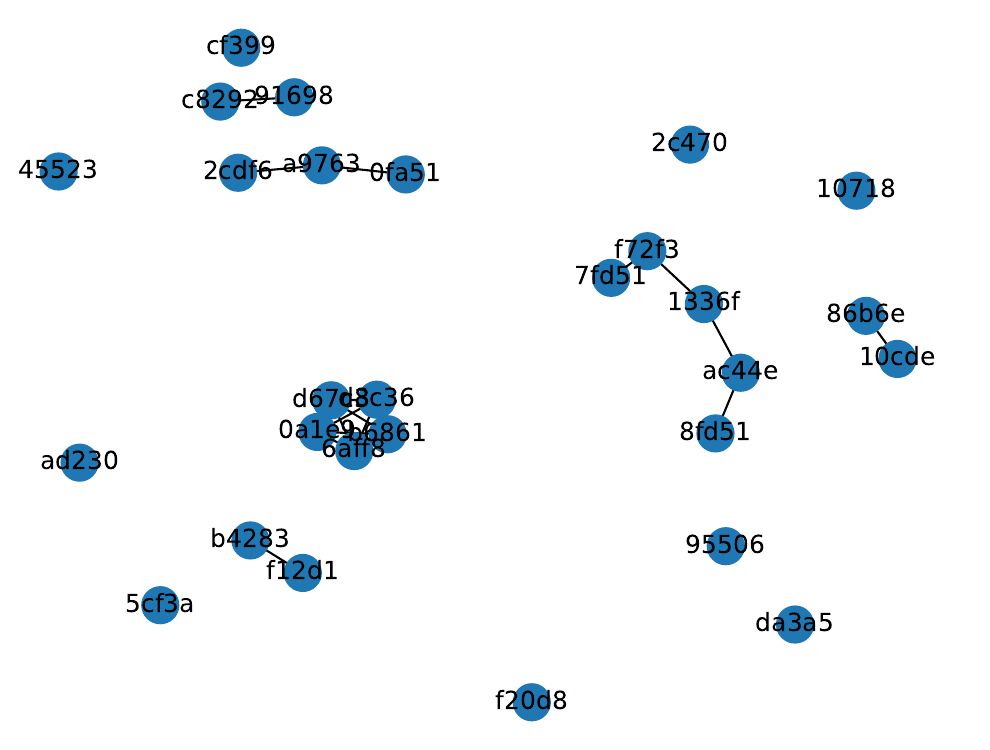}
  \caption{Friend graph for dataset {\classB}. On top of nodes are the anonymized student ids.}
  \Description{Friend graph for a university course where students have provided their friends, i.e. the people with whom they want to work together}
\label{fig:fgraph_519}
\end{figure}

\begin{figure}
  \centering
  \includegraphics[width=\linewidth]{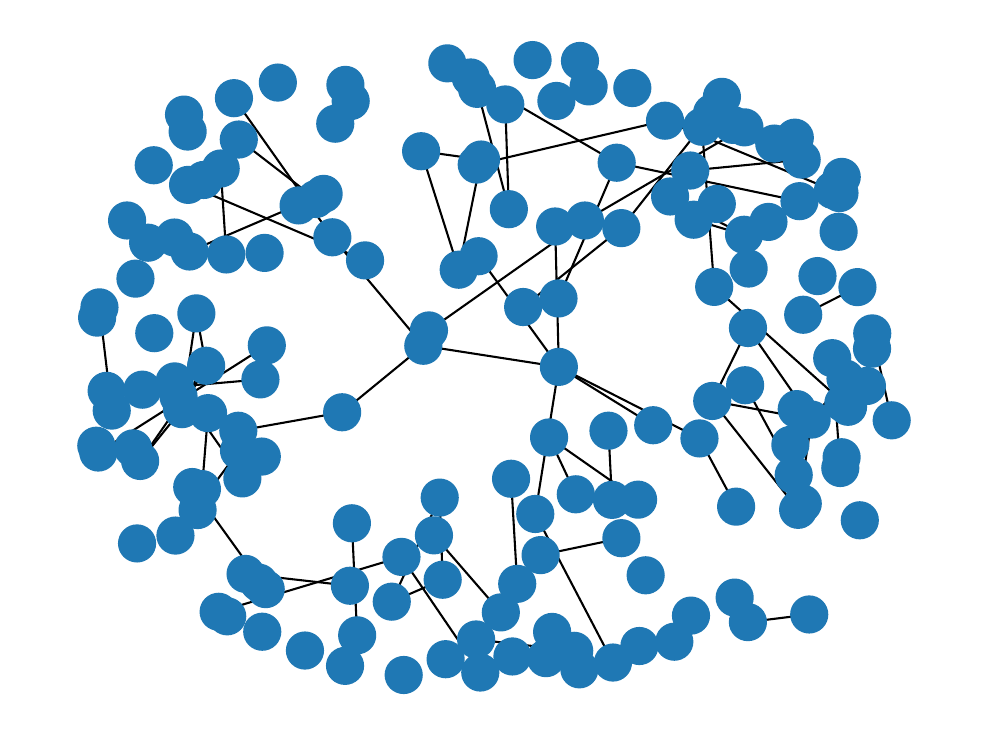}
  \caption{Friend graph for dataset {\classA}.}
  \Description{Friend graph for a university course where students have provided their friends, i.e. the people with whom they want to work together}
\label{fig:fgraph_506}
\end{figure}

Figures \ref{fig:fgraph_519} and \ref{fig:fgraph_506}
show the friend graphs (i.e., the complement of the conflict graphs) which is the input
in the education datasets. Observe the sparsity of both friend graphs.

\begin{figure}
  \centering
  \includegraphics[width=\linewidth]{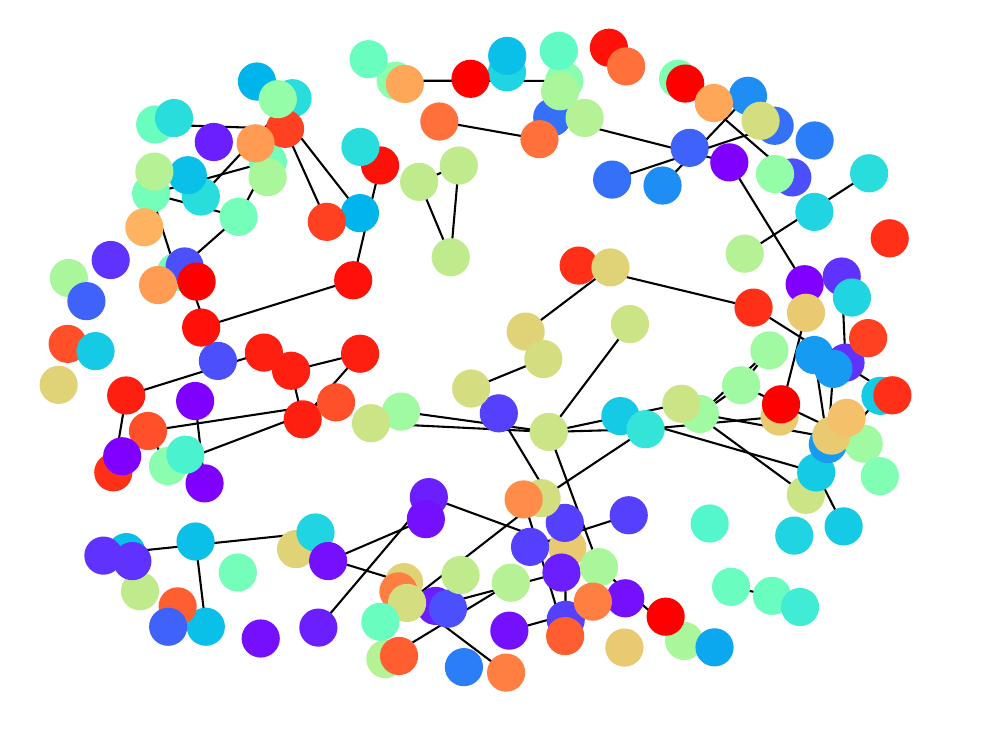}
  \caption{Friend graph colored using the optimal team assignment (\quadratic) for {\classA}. Labels are omitted for clarity.}
  \Description{Friend graph for a university course where students have provided their friends, i.e. the people with whom they want to work together. The nodes have been colored according to their assignment to teams.}
  \label{fig: friends-graph-A}
\end{figure}

\begin{figure}
  \centering
  \includegraphics[width=\linewidth]{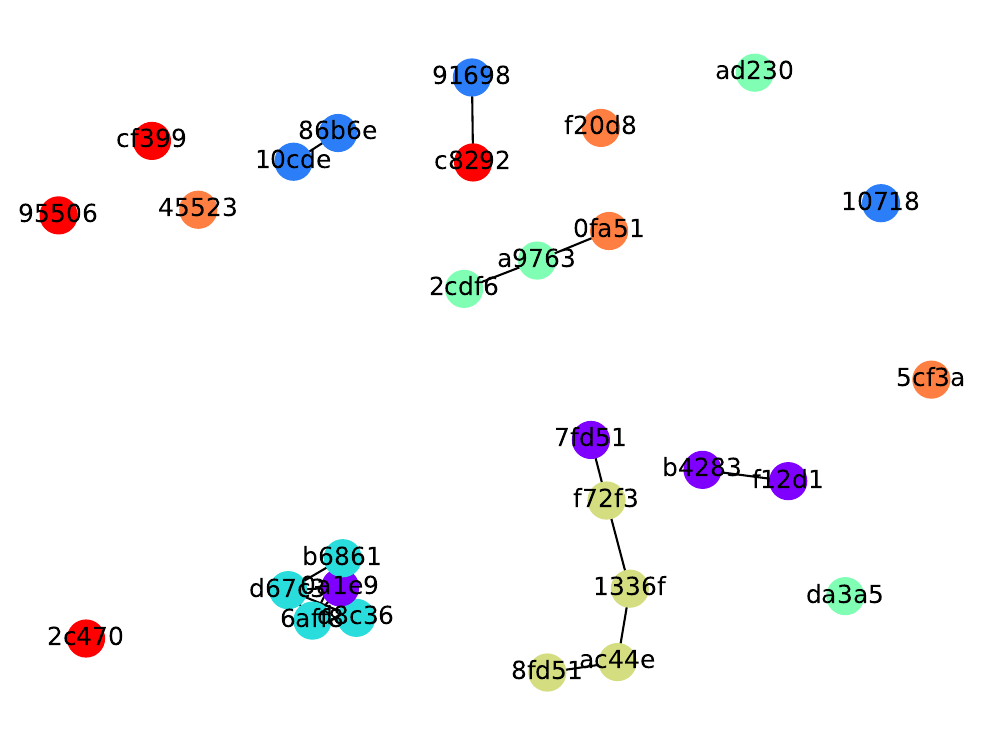}
  \caption{Friend graph colored using the optimal team assignment (\quadratic) for {\classB}. On top of nodes are the anonymized student ids.}
  \Description{Friend graph for a university course where students have provided their friends, i.e. the people with whom they want to work together. The nodes have been colored according to their assignment to teams.}
  \label{fig: friends-graph-B}
\end{figure}

Figure ~\ref{fig: friends-graph-A} and figure ~\ref{fig: friends-graph-B} show the friend graph for {\classA} and {\classB} respectively, where the nodes with the same color are assigned to the same project.

\subsection{Approximation ratios for {\linnorm} preference function} \label{appendix: linnorm-preference}
\begin{figure}%[H]
  \centering
  \includegraphics[width=\linewidth]{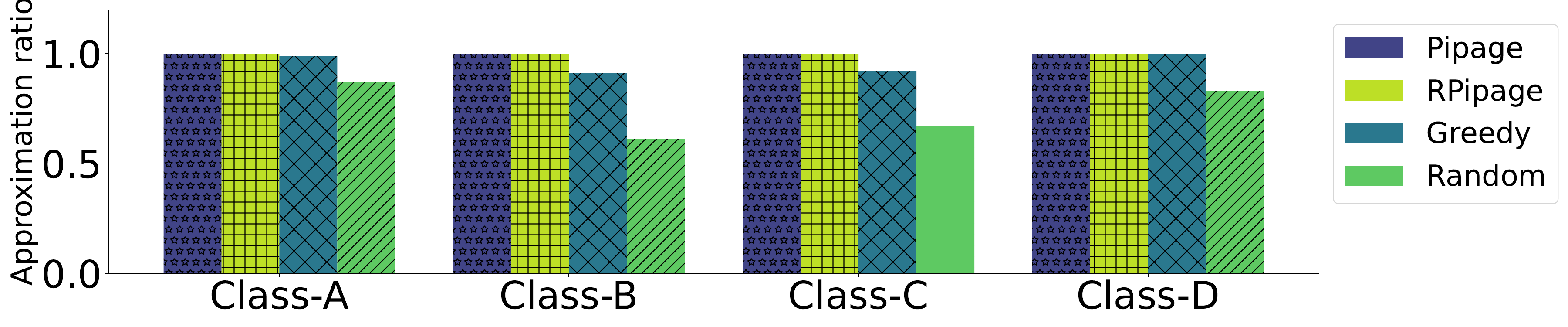}
  \caption{Education data; approximation ratios for education datasets. For all datasets we used the {\linnorm} project preference function and $\alpha = 10$.}
  \Description{Approximation ratios for education datasets}
  \label{fig: competitive-ratio-educational-linnorm}
\end{figure}

Figure ~\ref{fig: competitive-ratio-educational-linnorm} shows the approximation ratios for the {\linnorm} preference function.

\subsection{Hyperparameter tuning for the education datasets}\label{appendix:tuning-edu}
\begin{figure}%[h]
  \centering
  \includegraphics[width=\linewidth]{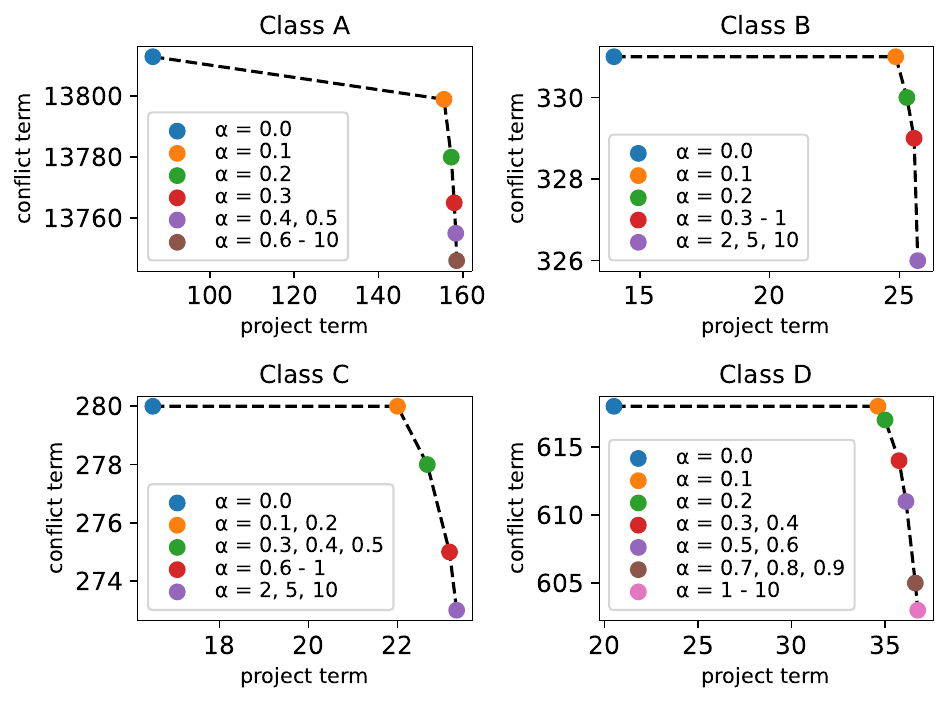}
  \caption{Education data; hyperparameter $\alpha$ tuning. We used the {\linnorm} project preference function.}
  \Description{Hyperparameter tuning}
  \label{fig:hyper-tuning-edu-linnorm}
\end{figure}

\begin{figure}%[H]
  \centering
  \includegraphics[width=\linewidth]{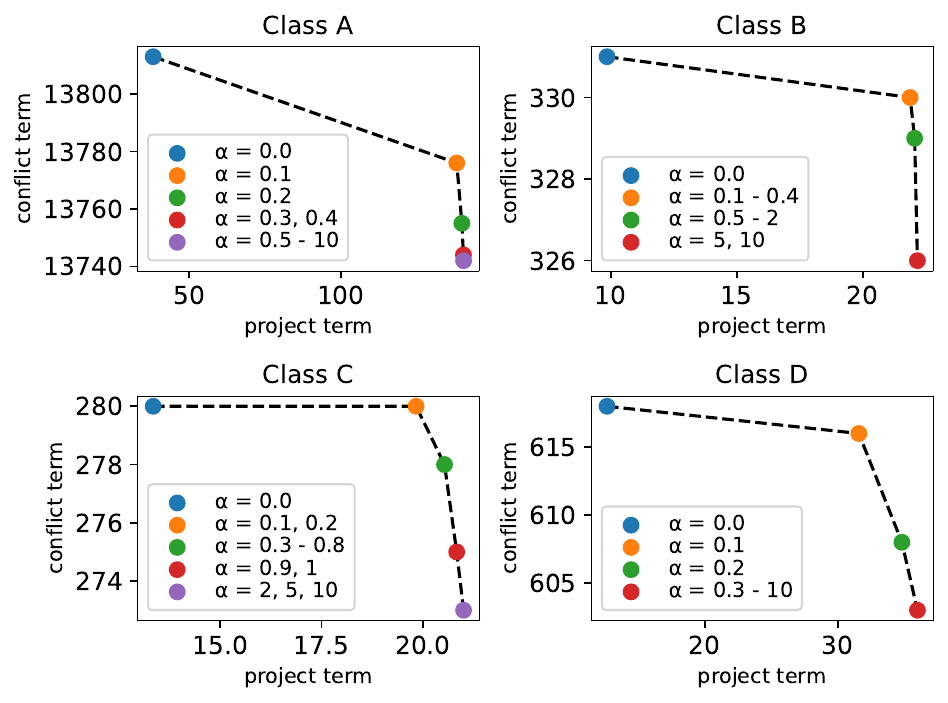}
  \caption{Education data; hyperparameter $\alpha$ tuning. We used the {\inverse} project preference function.}
  \Description{Hyperparameter tuning}
  \label{fig:hyper-tuning-edu-inverse}
\end{figure}

In Section~\ref{sec:lambda}, we gave a preview of how to tune the hyperparameter $\lambda$ or $\alpha$
by computing the terms $\left(F_\conflictgraph^\alpha, F_\preferencegraph^\alpha\right)$
for different values of $\alpha$. This results in an "elbow" plot that illustrates the trade-off between the two terms and allows us to tune the hyperparameter in a informed way. We present the plots for the {\linnorm} project
preference function in Figure ~\ref{fig:hyper-tuning-edu-linnorm}; 
the plots for the {\inverse} project preference function are very similar 
and are presented in Figure~\ref{fig:hyper-tuning-edu-inverse}.

Note that when $\alpha = 0$, the conflict term is almost equal to the number of conflict edges, i.e. almost all individuals in conflict are placed in different teams.
When $\alpha = 10$, the project preference term is almost equal to the number of individuals, i.e. almost all individuals get their highest-ranked project. This is because $c_{vt} = 1$, if the highest-ranked project of $v$ is $t$.

\subsection{Qualitative results for education datasets using the {\inverse} project preference function} \label{appendix: inverse-friends}

Table ~\ref{tab:stats_friends_inverse} shows that students got the same (or almost the same) average number of friends using the {\quadratic}, {\pipage} and {\randpip} algorithms as in the {\manual} assignment.

\begin{table}
\small
  \caption{Education data; $\mathcal{M}Q_\conflictgraph(\x_\mathcal{A})$  for $\mathcal{M}=\{\textit{max}, \textit{avg}\}$ of number of friends per student and $\mathcal{A}$ being all 
  algorithms; we also report $\textit{std}$ - the standard deviation for $\textit{avg}Q$. For all datasets we used the {\inverse} project preference function and $\alpha = 10$.}
  \label{tab:stats_friends_inverse}
  \begin{tabular}{ccccl}
    \toprule
    Dataset & Algorithm & ${\max}Q_\conflictgraph$ & $\textit{avg}Q_\conflictgraph$ & $\textit{std}$\\
    \midrule
    \multirow{6}{*}{\rotatebox{90}{{\classA}}} &          \quadratic &  4 & 0.65 & 0.78 \\
          & \pipage & 2 & 0.1 & 0.34 \\
          & \randpip & 1 & 0.08 & 0.26 \\
          & \greedy & 3 & 0.65 & 0.74 \\
          & \random & 1 & 0.4 & 0.49 \\
          & \manual & 3 & 0.63 & 0.63 \\
    \midrule
    \multirow{6}{*}{\rotatebox{90}{{\classB}}} &            {\quadratic} & 2 & 0.64 & 0.77 \\
            & {\pipage} & 2 & 0.64 & 0.77 \\
            & {\randpip} & 2 & 0.57 & 0.78 \\
            & {\greedy} & 3 & 0.79 & 1.08 \\
            & {\random} & 2 & 1.5 & 0.87 \\
            & {\manual} & 2 & 0.64 & 0.67 \\
    \midrule
    \multirow{6}{*}{\rotatebox{90}{{\classC}}} &          \quadratic & 2 & 0.54 & 0.57 \\
           & \pipage & 1 & 0.38 & 0.49 \\
           & \randpip & 1 & 0.38 & 0.49 \\
           & \greedy & 2 & 0.69 & 0.67 \\
           & \random & 1 & 0.54 & 0.5 \\
           & \manual & 2 & 0.69 & 0.54 \\
    \midrule
    \multirow{6}{*}{\rotatebox{90}{{\classD}}} &           \quadratic & 3 & 0.43 & 0.72 \\
        & \pipage & 1 & 0.29 & 0.45 \\
        & \randpip & 1 & 0.23 & 0.42 \\
        & \greedy & 2 & 0.38 & 0.54 \\
        & \random & 0 & 0.0 & 0.0 \\
  \bottomrule
\end{tabular}
\end{table}

\subsection{Qualitative results for education datasets using the {\linnorm} project preference function}

Tables ~\ref{tab:stats_projects_linnorm} and ~\ref{tab:stats_friends_linnorm} contain the qualitative results for the {\linnorm} project preference function.
Our algorithms are again comparable or better than the baselines. An observation is that using the {\linnorm} project preference function does not make a significance difference in the qualitative results.
Note again that our main focus on the education datasets was assigning students to projects they like. Assigning them with friends was a secondary goal.

\begin{table}
  \caption{Education data; $\mathcal{M}Q_\preferencegraph(\x_\mathcal{A})$  for $\mathcal{M}=\textit{max}, \textit{avg}$ of assigned project preferences and $\mathcal{A}$ being all 
  algorithms; we also report $\textit{std}$ - the standard deviation for $\textit{avg}Q_
  \preferencegraph$. For all datasets we used the {\linnorm} project preference function and $\alpha = 10$.}
  \label{tab:stats_projects_linnorm}
  \begin{tabular}{ccccl}
    \toprule
    Dataset & Algorithm & ${\max}Q_\preferencegraph$ & $\textit{avg}Q_\preferencegraph$ & $\textit{std}$\\
    \midrule
    \multirow{6}{*}{\rotatebox{90}{{\classA}}} &            \quadratic & 14.0 & 1.79 & 2.42 \\
        & \pipage & 14.0 & 1.81 & 2.47 \\
        & \randpip & 14.0 & 1.81 & 2.48 \\
        & \greedy & 14.0 & 1.95 & 2.50 \\
        & \random & 14.0 & 7.20 & 4.33 \\
        & \manual & 14.0 & 2.79 & 2.44 \\
    \midrule
    \multirow{6}{*}{\rotatebox{90}{{\classB}}} &            \quadratic & 4.0 & 1.57 & 0.86 \\
        & \pipage & 4.0 & 1.57 & 0.90 \\
        & \randpip & 4.0 & 1.57 & 0.90 \\
        & \greedy & 7.0 & 2.18 & 1.79 \\
        & \random & 7.0 & 4.36 & 2.24 \\
        & \manual & 3.0 & 1.71 & 0.80 \\
    \midrule
    \multirow{6}{*}{\rotatebox{90}{{\classC}}} &            \quadratic & 6.0 & 1.62 & 1.11 \\
        & \pipage & 6.0 & 1.62 & 1.15 \\
        & \randpip & 6.0 & 1.62 & 1.15 \\
        & \greedy & 6.0 & 2.12 & 1.48 \\
        & \random & 6.0 & 3.54 & 1.67 \\
        & \manual & 6.0 & 2.04 & 1.09 \\
    \midrule
    \multirow{6}{*}{\rotatebox{90}{{\classD}}} &            \quadratic & 1.875 & 1.05 & 0.20 \\
        & \pipage & 1.875 & 1.05 & 0.20 \\
        & \randpip & 1.875 & 1.05 & 0.20\\
        & \greedy & 1.875 & 1.05 & 0.20 \\
        & \random & 7.125 & 4.17 & 1.92 \\
  \bottomrule
\end{tabular}
\end{table}

\begin{table}
  \caption{Education data; $\mathcal{M}Q_\conflictgraph(\x_\mathcal{A})$  for $\mathcal{M}=\{\textit{max}, \textit{avg}\}$ of number of friends per student and $\mathcal{A}$ being all 
  algorithms; we also report $\textit{std}$ - the standard deviation for $\textit{avg}Q$. For all datasets we used the {\linnorm} project preference function and $\alpha = 10$.}
  \label{tab:stats_friends_linnorm}
  \begin{tabular}{ccccl}
    \toprule
    Dataset & Algorithm & ${\max}Q_\conflictgraph$ & $\textit{avg}Q_\conflictgraph$ & $\textit{std}$\\
    \midrule
    \multirow{6}{*}{\rotatebox{90}{{\classA}}} &           \quadratic & 3 & 0.71 & 0.79 \\
        & \pipage & 2 & 0.15 & 0.37 \\
        & \randpip & 1 & 0.10 & 0.30 \\
        & \greedy & 3 & 0.65 & 0.74 \\
        & \random & 0 & 0.00 & 0.00 \\
        & \manual & 3 & 0.63 & 0.63 \\
    \midrule
    \multirow{6}{*}{\rotatebox{90}{{\classB}}} &             \quadratic & 2 & 0.64 & 0.77 \\
        & \pipage & 2 & 0.5 & 0.63 \\
        & \randpip & 2 & 0.43 & 0.62 \\
        & \greedy & 3 & 0.79 & 1.08 \\
        & \random & 2 & 1.5 & 0.87 \\
        & \manual & 2 & 0.64 & 0.67 \\
    \midrule
    \multirow{6}{*}{\rotatebox{90}{{\classC}}} &           \quadratic & 1 & 0.54 & 0.50 \\
        & \pipage & 1 & 0.38 & 0.49 \\
        & \randpip & 1 & 0.46 & 0.50 \\
        & \greedy & 2 & 0.69 & 0.67 \\
        & \random & 1 & 0.46 & 0.50 \\
        & \manual & 2 & 0.69 & 0.54 \\
    \midrule
    \multirow{6}{*}{\rotatebox{90}{{\classD}}} &            \quadratic & 3 & 0.43 & 0.72 \\
    & \pipage & 1 & 0.23 & 0.42 \\
    & \randpip & 1 & 0.29 & 0.45 \\
    & \greedy & 2 & 0.38 & 0.54 \\
    & \random & 0 & 0.00 & 0.00 \\
  \bottomrule
\end{tabular}
\end{table}

\subsection{Hyperparameter tuning for the {\diversity} dataset}\label{appendix:tuning-employee}

\begin{figure}
  \centering
    \includegraphics[width=\linewidth]{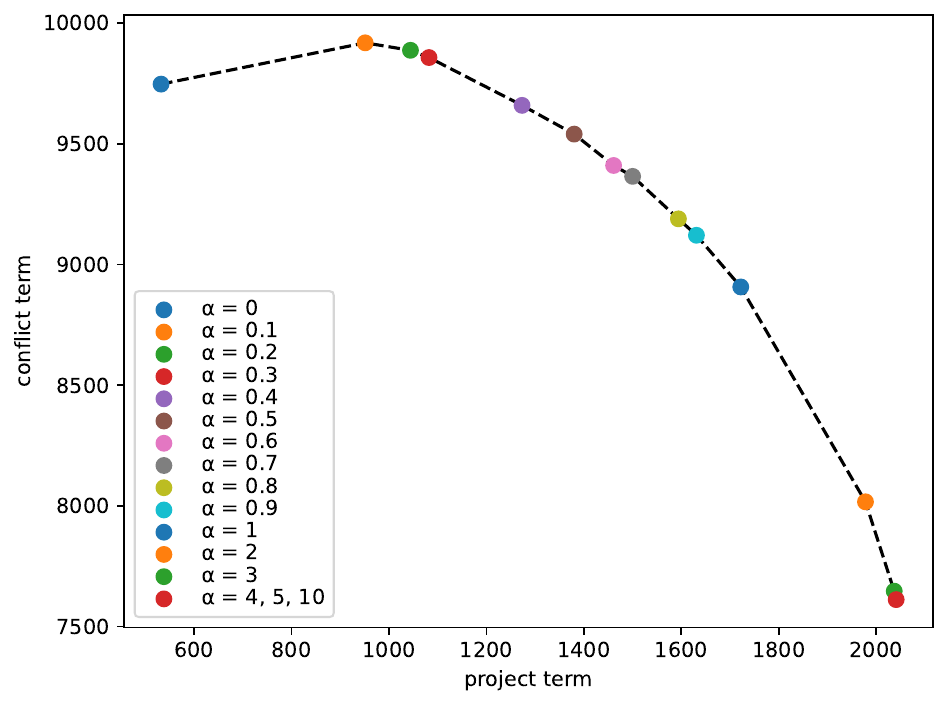}
  \caption{Employee data; Hyperparameter $\alpha$ tuning for the {\diversity} dataset.}
  \Description{Hyperparameter tuning for the diversity dataset}
  \label{fig:hyperparameter-diversity-objective}
\end{figure}

% \begin{figure}
%   \centering    
%   \includegraphics[width=\linewidth]{images/hyperparameter_diversity_objective.png}
%   \caption{Employee data; Controlling the balance between the fraction of people who changed department \textit{PER} and average gender gap per department \textit{AVG-GAP} using $\alpha$.}
%   \Description{Hyperparameter tuning for the diversity dataset. Balancing the social satisfaction terms}
%   \label{fig:hyperparameter-diversity-objective}
% \end{figure}

\begin{figure}
  \centering    
  \includegraphics[width=\linewidth]{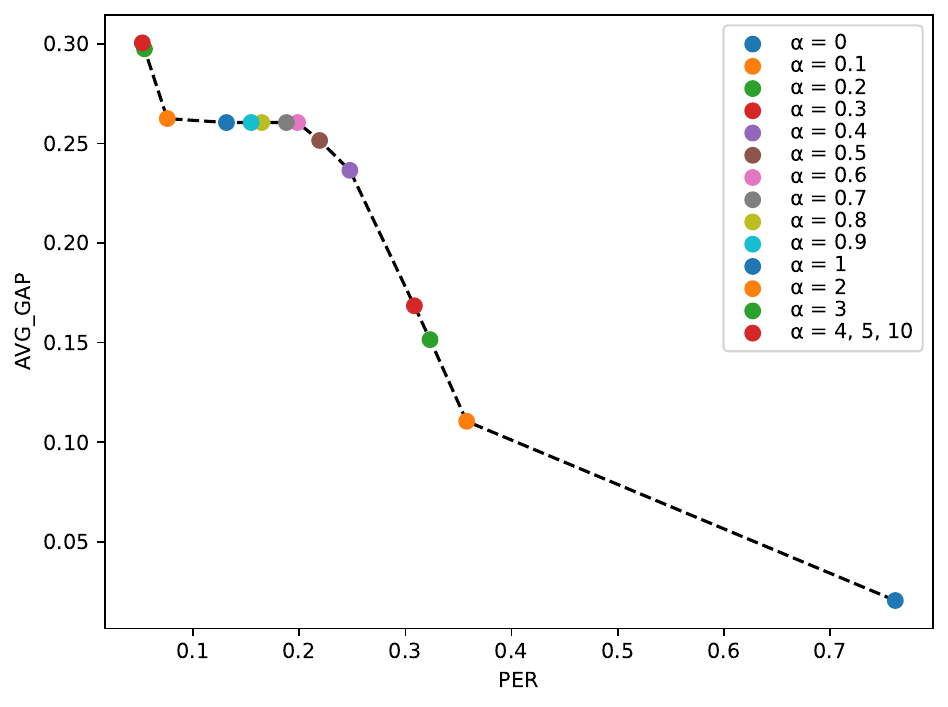}
  \caption{Employee data; Controlling the balance between the fraction of people who changed department \textit{PER} and average gender gap per department \textit{AVG-GAP} using $\alpha$.}
  \Description{Hyperparameter tuning for the diversity dataset}
  \label{fig:hyperparameter-diversity}
\end{figure}

Figure ~\ref{fig:hyperparameter-diversity} shows the trade-off between the percentage of people who changed department and the average percentage of the male-female gap per department. Note that for values of $\alpha$ between $0.5$ and $2$ there is a plateau. That is, the average gender gap per department remains almost constant although more employees change department. After $\alpha$ drops below $0.5$ the average male-female gap drops significantly, but the percentage of people who change departments grows very fast. According to the above plot a reasonable choice for $\alpha$ is $\alpha = 2$.

Figure ~\ref{fig:hyperparameter-diversity-objective} shows the trade-off between the task and social satisfaction terms.

\end{document}